\documentclass[preprint]{elsarticle}
\usepackage{hyperref}
\usepackage[centertags]{amsmath}
\usepackage{amsfonts}
\usepackage{amssymb}
\usepackage{adjustbox}
\usepackage{amsmath, bm}
\usepackage[ruled]{algorithm2e}
\usepackage{enumitem}
\usepackage{multicol}
\usepackage{multirow}
\usepackage{CJKutf8}
\usepackage{xcolor}
\usepackage{hyperref}
\usepackage{changes}

\usepackage{makecell}
\usepackage{ragged2e}
\usepackage{changes}
\usepackage{color}
\usepackage{ulem}

\newtheorem{remark}{Remark}
\journal{Journal}
\numberwithin{table}{section}
\usepackage[ruled]{algorithm2e}
\SetKwBlock{GrO}{Grouping optimization:}{end}
\SetKwBlock{PeO}{Perform optimization steps:}{end}

\numberwithin{equation}{section}
\numberwithin{figure}{section}
\begin{document}

\newcommand{\hei}{\CJKfamily{hei}}

\begin{CJK}{UTF8}{gbsn}
\begin{frontmatter}
\title{Hard constraint learning approaches with trainable influence functions for evolutionary equations}
\author[1,2,3]{Yushi Zhang} 
\author[4]{Shuai Su\corref{cor1}}
\ead{shuaisu@bjut.edu.cn}
\author[2]{Yong Wang}
\author[1,3]{Yanzhong Yao\corref{cor1}}
\ead{yao\_yanzhong@iapcm.ac.cn}

\cortext[cor1]{Corresponding author.}

\address[1]{Institute of Applied Physics and Computational
Mathematics, Beijing 100088, China}
\address[2]{Graduate School of China Academy of Engineering Physics, Beijing 100088, China}
\address[3]{National Key Laboratory of Computational Physics, Beijing 100088, China}
\address[4]{School of Mathematics, Statistics and Mechanics, Beijing University of Technology, Beijing 100124, China}

\begin{abstract}


This paper develops a novel deep learning approach for solving evolutionary  equations, which integrates sequential learning   strategies  with an enhanced hard constraint strategy featuring trainable parameters, addressing the low computational accuracy of standard Physics-Informed Neural Networks (PINNs) in  large temporal domains. 
 Sequential learning strategies divide   a large  temporal domain  into multiple  {subintervals }  and  solve  them one by one in a chronological order, which naturally  respects the principle of causality  and improves the stability of the PINN solution. The improved hard constraint strategy  strictly ensures the continuity and smoothness of the PINN solution at time interval nodes, and at the same time passes the  {information} from the previous interval to the next interval, which avoids the incorrect/trivial solution at the position far from the initial time. Furthermore, by investigating  {the requirements of different types of equations} on hard constraints, we design a novel influence function with trainable parameters for hard constraints, which provides theoretical and technical support for the  {effective} implementation{s} of hard constraint strategies, and significantly improves the universality and computational accuracy of our method. In addition, an adaptive time-domain partitioning algorithm is
proposed, which plays an important role in the application of the proposed method as well as in the improvement of computational efficiency and accuracy. Numerical experiments verify the   performance of the method. 
 {The data and code accompanying this paper are available at \href{https://github.com/zhizhi4452/HCS}{https://github.com/zhizhi4452/HCS}.}
\end{abstract}

\end{frontmatter}

\section{Introduction}
Evolutionary equations are a class of partial differential equations (PDEs) that formulate the evolution of physical phenomena over time and are widely used in many fields. 
In electromagnetism and hydrodynamics, the wave equation can reflect the dynamic behavior of electromagnetic and water waves. 
In thermodynamics, the heat conduction equation describes the law of heat propagation in different media. 
In quantum mechanics, the
Schrodinger equation describes the evolution of quantum states. 
All of the above are time-dependent equations.

Physics-Informed Neural Networks(PINNs) \cite{712178,RAISSI2019686}   {are completely new tools}  for solving PDEs that combine deep learning and the laws of physics. PINNs show  the advantages in solving PDEs, especially when dealing with nonlinear and high-dimensional problems \cite{Karniadakis}. However, when solving evolutionary equations over large time domain,  standard PINNs face the following two challenges:
\begin{enumerate}
    \item 
In the case where the boundary conditions do not provide labeled data, e.g. {Symmetric or Neumann}   boundary conditions do not explicitly provide the exact  solution  at the boundaries of the domain, PINNs tends to give low accuracy, even incorrect predictions, as the temporal domain  becomes large~\cite{SOIBAM2024125480,PENWARDEN2023112464}.
The main reason lies in the fact that the label{ed} data provided by the governing equations are merely the initial conditions. For the computational domain far from the initial time, since the optimization algorithm does not  {adhere} the temporal causality during the optimization process, the neural network  {may shed} the influence of the initial conditions during training and be solely constrained by the PDEs. Theoretically, the solutions that merely satisfy the PDEs are not unique, which in turn leads to  {inaccurate  predictions} of the neural network over long temporal domain.

    \item 
 
Over a
long  temporal domain, evolutionary equations often exhibit multiscale characteristics, with physical quantities or their rates of change varying significantly at different times. To accurately capture such multiscale phenomena,  {it is necessary} to enhance the depth or width of neural networks. This not only significantly increases computational   load but, more importantly, 
  makes training extremely challenging  due to the significant increase in the problem's inherent complexity. Consequently, higher demands  {are placed on the arrangement of} training points, the selection of neural network architectures, and the types of optimization algorithms and  {hyperparameter settings} employed~\cite{Wang2020OnTE,WANG2024113112}.

\end{enumerate}

In response to the two aforementioned challenges faced by  {standard}  PINNs when solving evolutionary equations over  large temporal domains, researchers have proposed various methods. Wang et al. \cite{Wang2024RespectingCF} constructed an modified loss function that ensures the training  of neural network 
respects the principle of causality, with the optimization process proceeding sequentially over time. This method significantly enhances the computational accuracy of evolutionary equations and prevents the emergence of trivial solutions. However, the authors also noted that this method entails excessive computational overhead. Wang et al. \cite{WANG2024106998}, leveraging the extrapolation capabilities of PINNs,  {designed}   an extrapolation-driven neural network architecture, providing an effective approach for solving evolutionary equations.
Researchers have extensively studied sequential learning  {strategies} to address the aforementioned two challenges. Sequential learning is a strategy that enables PINNs' training process to naturally adhere to temporal causality by partitioning the time domain into multiple intervals and training them sequentially to obtain solutions over the entire temporal domain. Wight and Zhao \cite{CiCP-29-930} applied this strategy to solve the phase-field model. Mattey and Ghosh \cite{MATTEY2022114474} improved the sequential learning strategy by using the training results from the previous time interval as  {extra supervised learning points}, further ensuring the enforcement of temporal causality. Guo et al. \cite{guo2023pretraining}, based on pre-training strategy and incorporating  {extra supervised learning points}, provided an efficient and highly accurate PINN version for evolutionary equations. Penwarden et al. \cite{PENWARDEN2023112464} summarized various sequential learning methods and proposed causal sweeping strategies that adheres to causality. Jung et al. \cite{JUNG2024117036} constructed integral loss functions on each interval and trained sequentially from the initial time, thereby seamlessly integrating temporal causality into the training process. All of the above sequential learning methods have significantly improved the computational accuracy of standard PINNs for evolutionary equations.

It needs to be emphasized that existing sequential learning strategies employing multiple neural networks mostly fail to strictly satisfy the continuity and smoothness of the exact solution with respect to the time variable at partition nodes. To address this shortcoming, Roy and Castonguay \cite{Exact} proposed an ingenious method called HCS-PINN (Hard Constrained Sequential PINN), which ensures the smoothness of PINN solutions at partition nodes during the sequential learning process. However, to achieve higher accuracy, they had to employ some specialized techniques, such as hard periodic boundary conditions and empirical causal weights, which are often problem-specific and lack generality.

To address the stability and accuracy issues  of PINNs when solving evolutionary equations over a large temporal  domain, we construct \textbf{\textit{constraint influence functions with trainable parameters}} at interval nodes and design a rich set of hard constraint formulations.  By integrating these with sequential learning strategies, we propose a robust   and high accuracy PINN method  for solving evolutionary equations over a large time domain. Interestingly, the HCS-PINN proposed in \cite{Exact} derived formulas approximately similar to ours from a different perspective.
However, our analytical approach and derivation  differ from theirs. The design of our method fully considers the properties of the governing equations, making the proposed method more general and capable of adaptively solving different types of governing equations.  Our method can be considered an extended version of HCS-PINN, with HCS-PINN being a special case of our method.

The main contributions of this paper are as follows: 
\begin{itemize}
 \item 
By dividing the entire time domain into multiple subintervals and training them sequentially, this method naturally satisfies the temporal causality during the training process, transforming the challenging problem over a large temporal domain  into multiple simpler sub-problems, and ensuring the efficient acquisition of accurate predictive functions across the entire temporal  domain.

 \item By employing hard constraint techniques, the  predictive results of the neural network are able to strictly maintain the continuity and smoothness  inherent in the exact solution at interval nodes. This approach also  {ingeniously passes information} from the previous time interval to the next, avoiding incorrect or trivial solutions in regions far from the initial values due to the loss of influence from initial conditions.

 \item  
By setting constraint influence functions, a practical strategy is provided for the effective implementation of hard constraint techniques. This enhances the mechanistic understanding of hard constraint techniques and develops its usage methods. This strategy not only enhances the understanding of the generalization ability of PINNs but also effectively improves prediction accuracy over a long time domain.

  \item  
By constructing a posteriori approximations of the $L_2$ relative error, an adaptive partitioning algorithm is provided for dividing the entire temporal domain  into multiple intervals, addressing the issue of reasonable partitioning of time domain. This not only benefits the improvement of PINN solution accuracy across the entire domain but also enhances computational efficiency. To our knowledge, this is  {the first}  adaptive temporal domain  partitioning algorithm presented in sequential learning strategies.
 
\end{itemize}

The remainder of this paper is organized as follows. In Section \ref{sec2},  a sequential PINN method is discussed based on an improved hard constraint strategy. This section first presents a hard constraint method for evolutionary equations to strictly satisfy the initial conditions, then introduces an improved hard constraint strategy by incorporating constraint influence functions, and finally applies the advanced hard constraint strategy to sequential learning,   to design  a novel PINN method for  evolutionary equations. In Section \ref{sec3}, an algorithm for   reasonable partitioning of the entire time domain is proposed. In Section \ref{sec4}, numerical examples are provided to validate the performance of the proposed method. Finally, a conclusion of the work is given in 
 Section \ref{sec5}.

\section{A novel hard constraint strategy-based PINN method for solving evolutionary equations}\label{sec2}

\subsection{Preliminaries}

\subsubsection{ Standard PINN method}

Consider the  general form of
evolutionary equations as follows
\begin{equation}\label{time-dependent PDE}
\begin{cases}
\frac{\displaystyle{\partial u}}{\displaystyle{\partial t}} + \mathcal{P}(u)=0,&x\in \Omega \subset \mathbb{R}^{d},t\in(T_0,T],\\
u(x,T_0)=\mathcal{I}(x),&x\in \Omega,\\
\mathcal{B}(u, x, t)=0,&x\in \partial \Omega, t\in(T_0,T],\\
\end{cases}
\end{equation}
where $u=u(x,t)$ is the solution on the spatio-temporal domain $\Omega \times (T_0, T]$, 
$\mathcal{P}$ denotes the differential operator   with respect to the spatial variable $x$, $\mathcal{I}(x)$ is the initial  {condition}  and $\mathcal{B}(u,x,t)=0$ represents the general form of the boundary conditions, and the types of boundary conditions can include Dirichlet, Neumann, Robin, Periodic, and Mixed types, among others.

Following the PINN framework proposed in \cite{RAISSI1}, the solution to   \eqref{time-dependent PDE} is given by a prediction function $u_\theta(x,t)$ expressed by a neural network, obtained by optimizing the following loss function:
\begin{equation}\label{loss_std}
\begin{cases}
\mathcal{L}(\theta)=w_i\mathcal{L}_i(\theta;\tau_i)+w_b\mathcal{L}_b(\theta;\tau_b)+w_r\mathcal{L}_r(\theta;\tau_r),\\
\mathcal{L}_i(\theta;\tau_{i})=\frac{1}{N_0} \sum_{i=1}^{N_0}\left| u_\theta \left(x_i,T_0\right)-\mathcal{I}(x_i)  \right|^2, \\
\mathcal{L}_b(\theta;\tau_{b})= \frac{1}{N_b} \sum_{i=1}^{N_b}\left|\mathcal{B}(  u_\theta(x_i,t_i),x_i,t_i)\right|^2 , \\
\mathcal{L}_r(\theta;\tau_{r})=\frac{1}{N_r} \sum_{i=1}^{N_r}\left| \frac{\displaystyle{\partial u_{\theta}}}{\displaystyle{\partial t}}(x_i,t_i) +  \mathcal{P}(u_\theta(x_i,t_i)) \right|^2.
\end{cases}
\end{equation}
Here,  
$\theta$ represents a set of neural network parameters to be optimized, $w_i$, $w_r$ and $w_r$
  are the weights for the initial loss term 
  $\mathcal{L}_i$, boundary loss term $\mathcal{L}_b$ and residual loss term $\mathcal{L}_r$, respectively, where $\mathcal{L}_i$ and $\mathcal{L}_b$
  are collectively referred to as supervised loss terms, $\tau_i$, $\tau_b$ and ${\tau_r}$
  represent the sets of initial  {sample}   points, boundary  {sample}  points, and residual  {sample} points, respectively,
  {
   $N_0$, $N_b$ and $N_r$  denote the size of $\tau_i$, $\tau_b$ and ${\tau_r}$, respectively.}

  Assuming the optimization of the loss function $\mathcal{L}(\theta)$, we obtain a set of network parameters \begin{equation}
 \hat{{\boldsymbol{\theta}}} = \arg\min_{{\boldsymbol{\theta}}} \mathcal{L}(\theta), 
\end{equation}
Then, the distribution function  
$u_{\hat\theta}(x,t)$ defined by the optimized neural network serves as a  potential  solution to   \eqref{time-dependent PDE}, which we refer to as the PINN solution.

\textit{ \eqref{time-dependent PDE} only includes the first-order time derivative $u_t$. The method discussed in this paper is also applicable to equations with higher-order time derivatives, such as the wave equation.}

\subsubsection{A hard-constrained PINN method that strictly satisfies the initial conditions}

The hard constraint technique enforces PINN solutions to strictly satisfy the  {definite}    conditions. Compared to  soft constraints, hard constraints eliminate the need to consider supervised terms in the loss function, avoiding the  {competition}   between supervised and residual terms. There is extensive research on hard constraints, see \cite{712178,article,Projection,Exact,PhyCRNet}. They incorporate the  {definite}  conditions 
into   the solution expressions of the equations, ensuring that the solutions strictly satisfy  the {definite}  conditions.

For boundary conditions, due to their diverse types and the complex shapes of boundaries in high-dimensional spaces, it is challenging to establish a unified paradigm for  hard constraints to boundary conditions, making their implementation difficult.

For initial conditions, it is straightforward to provide a general hard constraint formulation. A  simple one is as follows
\begin{equation}\label{eq:HC_YYZ1}
     u_H(x,t) = \mathcal{I}(x) + (t-T_0)\cdot u_{\boldsymbol{\theta}}(x,t), \quad t\in [T_0,T],\ x\in \bar\Omega.
\end{equation} 
Clearly, if $t=T_0$, then $u_H = \mathcal{I}(x)$, which is precisely the initial condition required by \eqref{time-dependent PDE}. Using the hard constraint above, the loss function is transformed into  
\begin{equation}\label{loss_HDTime}
\begin{cases}
\mathcal{L}_H(\theta)=w_b\mathcal{L}_b(\theta;\tau_b)+w_r\mathcal{L}_r(\theta;\tau_r),\\
\mathcal{L}_b(\theta;\tau_{b})= \frac{1}{N_b} \sum_{i=1}^{N_b}\left|\mathcal{B}(  u_H(x_i,t_i),x_i,t_i)\right|^2 , \\
\mathcal{L}_r(\theta;\tau_{r})=\frac{1}{N_r} \sum_{i=1}^{N_r}\left| \frac{\displaystyle{\partial u_H}}{\displaystyle{\partial t}}(x_i,t_i) +  \mathcal{P}(u_H(x_i,t_i)) \right|^2.
\end{cases}
\end{equation}
Here, the loss function $\mathcal{L}_H(\theta)$ compared to $\mathcal{L}(\theta)$ in  \eqref{loss_std}, omits the initial loss term.
  
Assuming the optimization of the loss function $\mathcal{L}_H(\theta)$ yields the network parameters 
\begin{equation}
 \hat{{\boldsymbol{\theta}}} = \arg\min_{{\boldsymbol{\theta}}} \mathcal{L}_H(\theta), 
\end{equation}
we can obtain the PINN solution  with the initial conditions serving as hard constraints
 \begin{equation}\label{eq:HC_PINN}
     \hat u_H(x,t) = \mathcal{I}(x) + (t-T_0)\cdot u_{\hat{\boldsymbol{\theta}}}(x,t), \quad t\in [T_0,T].
\end{equation}

  This paper denotes variables or functions after training by placing the symbol  $\ \hat{}$ 
  above their names.

  \subsection{Hard constraint formulation for initial conditions with influence functions}

The hard constraint   \eqref{eq:HC_YYZ1}  ensures that the PINN solution strictly satisfies the initial conditions at $t=T_0$. However, in practice, we find that this hard constraint   often leads to exceptionally difficult training of network parameters, sometimes even preventing the acquisition of effective PINN solutions.

Through analysis, we summarize the reasons for the poor practical performance of the hard constraint \eqref{eq:HC_YYZ1} as follows:

\begin{itemize}
    \item

In \eqref{eq:HC_YYZ1}, the influence of the initial   function $\mathcal{I}(x)$ on the PINN solution   always persists in a constant manner across the entire time domain. However, this constraint   lacks a mechanism to modulate the importance between the initial  function and the function to be determined. We know that the energy norm and maximum norm of the equation's solution are not only related to the initial condition but also to the source terms and boundary conditions. The  influence degree of the initial condition on the solution is not constant over time.

     \item 
     
     In many practical applications, the initial   function $\mathcal{I}(x)$ and the   solution function $u(x,t)$ differ in smoothness, potentially belonging to different function spaces. For instance, in the heat conduction equation, the initial function can be $\mathcal{I}(x)\in C^0(\Omega)$, while the   solution function  within the computational domain is $u(x,t)\in C^{2,1}(\Omega, t>T_0)$. When $\mathcal{I}(x)$ and $u(x,t)$ have different smoothness,  \eqref{eq:HC_YYZ1} is equivalent to expressing a smooth function with a non-smooth one.

\end{itemize}

Based on the first reason mentioned above, we make the following improvements to the hard constraint \eqref{eq:HC_YYZ1}: 
 \begin{equation}\label{eq:HC_YYZ2}
     u_H(x,t) = \lambda(t)\cdot\mathcal{I}(x) + \eta(t)\cdot u_{\boldsymbol{\theta}}(x,t),\quad t\in [T_0,T],
\end{equation}
where $\lambda(t)$ is used to characterize the influence degree of the initial function on the PINN solution at different times, which we refer to as the \textbf{\textit{influence function}} of the initial condition. $\eta(t)$ is a monitor function set to facilitate the implementation of hard constraint condition, which we refer to as the \textbf{\textit{adjoint function}} of $\lambda(t)$. These two functions are required to satisfy the following conditions:
\begin{enumerate}
    \item 
    $ 0 \leq \lambda(t) \leq 1, ~\lambda(T_0)=1$,\quad $0 \leq \eta(t) \leq 1, ~\eta(T_0)=0.$\\
    This condition  ensures that  the PINN solution strictly satisfies the initial condition at   $T=T_0$.

    \item 
    $\lambda^{'}(t) < 0, ~\eta^{'}(t) > 0,\quad t\in[T_0, T]$.\\
    This condition implies that   the influence  of the initial function on the PINN solution gradually decreases as time progresses.
\end{enumerate}

Based on the first condition, we typically choose $\lambda(t)=1-\eta(t)$.


\begin{remark}\label{remark1}
As mentioned above,   if the initial function $\mathcal{I}(x)$ and the   solution function $u(x,t)$ differ in smoothness, then using hard constraints often makes training difficult. We believe that, for this scenario, a soft constraint method that optimizes the traditional loss function \eqref{loss_std} should be employed.
\end{remark}

\subsection{ Temporal domain segmentation learning strategy based on novel hard constraint}

  Addressing the issue of low computational accuracy of standard PINN methods over a long temporal domain, this subsection presents a new sequential deep learning method. The method decomposes the entire temporal domain into several small intervals and then solves the governing equations on each interval using PINNs in conjunction with an novel hard constraint strategy.

  \subsubsection{
  Hard constraint learning strategy for partitioning the temporal domain  into  two intervals}
\label{sec:2sub}

First, we consider the case of dividing the entire temporal domain  $[T_0, T]$ into two intervals, i.e.,
 $$[T_0, T]=[T_0, T_1]\bigcup[T_1,T_2],\quad T_2=T.$$
In the first interval $[T_0, T_1]$, to avoid the issue of differing smoothness between the initial function and the solution function mentioned in Remark \ref{remark1}, we obtain the PINN solution for this interval using the traditional soft constraint PINN method, denoted as $u_{\hat\theta_1}(x,t)$. Compared to larger domains, available PINN solution can be relatively easily obtained on smaller time domains \cite{haitsiukevich2023improved}.

Now, we investigate the   method for the second interval $[T_1, T_2]$. In fact, we simply take the value of  $u_{\hat\theta_1}(x,t)$ at $t=T_1$, which is  
$u_{\hat\theta_1}(x,T_1)$, as the initial condition, and then apply the traditional soft constraint PINN method to obtain the PINN solution on the interval $[T_1, T_2]$. This is the approach adopted by many sequential methods \cite{PENWARDEN2023112464,CiCP-29-930,JUNG2024117036}. A notable drawback of this approach is that at the interval nodes, the PINN solution expressed by two separate neural networks fail to maintain the original continuity and smoothness of the exact solution of the equation.

Since $u_{\hat\theta_1}(x,t)$ and the solution over the interval $[T_1,T_2]$ are solutions of the same governing equation on two intervals,   {they possess identical smoothness properties.} Therefore, we can fully adopt a hard constraint strategy to solve for this interval. For the interval $[T_1,T_2]$, we construct the following hard constraint formula
\begin{equation}\label{eq:HC_YYZ3}
     u_{H_2}(x,t) = \lambda_2(t)\cdot u_{\hat\theta_1}(x,t) +  \cdot\eta_2(t)\cdot u_{\theta_2}(x,t),\quad t\in [T_1,T_2],
\end{equation}
where the subscript $2$ denotes the hard constraint   for the second interval. 

It is important to note that in \eqref{eq:HC_YYZ3}, only   $u_{\theta_2}(x,t)$ requires training, while $u_{\hat\theta_1}(x,t)$ is  known.

\begin{figure}[htbp]
    \centering
\includegraphics[width = 0.9\textwidth]{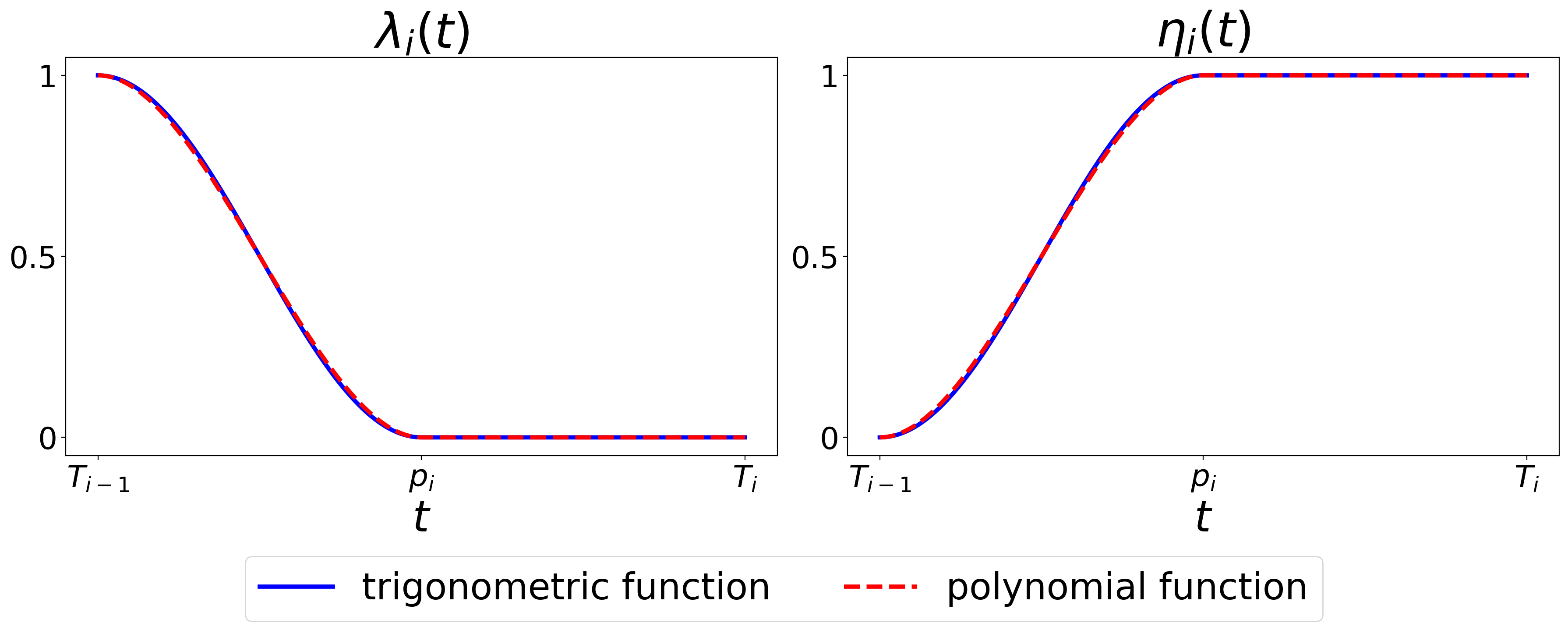}
\caption{Influence function  and  adjoint  function   using  different types of function.}
\label{fig:lambda-eta}
\end{figure}

 Considering that at the interval node $T_1$,   the PINN solution across the two intervals  should satisfy the required continuity and smoothness with respect to the time variable, i.e.,
\begin{equation}\label{continus}
{u}_{H_2}(x,T_1) = {u}_{\hat\theta_1}(x,T_1),
\end{equation}
\begin{equation}\label{d_continus}
\frac{\partial u_{H_2}(x,t)}{\partial t} \bigg|_{t=T_1} = \frac{\partial u_{\hat\theta_1}(x,t)}{\partial t} \bigg|_{t=T_1}.
\end{equation}
To this end, we require that $\lambda_2(t)$ and $\eta_2(t)$ have sufficient smoothness and satisfy the following conditions 
\begin{enumerate}
    \item $\lambda_2(T_1)=1,\quad\eta_2(T_1) =0.$\\
    This condition ensures    the continuity of the PINN solution from the two intervals at the interval node   $T_1$.
 
    \item $\lambda_2'(T_1)=0,\quad \eta_2'(T_1)=0.$\\
    This condition ensures  the smoothness of the PINN solution from the two intervals at the interval node   $T_1$.
    
    \item $\lambda_2(T_2)=0,\quad\eta_2(T_2) =1$.\\
    This condition indicates that the PINN solution of the preceding interval directly affects only the current interval, thereby facilitating the extension 
 of the method from two intervals to multiple intervals. In fact, a more stringent condition should be
\begin{equation}\label{cond_yyz1}
       \lambda_2(t)=0,\quad\eta_2(t) =1, \quad t\in[T_2,+\infty).
    \end{equation}
    
    \item 
    $\lambda_2'(T_2) =0, \quad\eta_2'(T_2)=0.$\\
    This condition facilitates the  applicability of the method to multiple intervals. A more stringent condition should be  
\begin{equation}\label{cond_yyz2}
       \lambda_2'(t) =0, \quad\eta_2'(t)=0, \quad t\in[T_2,+\infty).
    \end{equation}
    
    \item $\lambda_2'(t)\le 0 ,\quad \eta_2'(t)\ge 0,\quad t\in(T_1,T_2).$\\
    This condition is not essential,  
    and it indicates that the influence of the initial condition $u_{\hat\theta_1}(x,T_1)$ 
gradually decreases over time, while  the weight of the part $u_{\theta_2}(x,t)$  gradually increases.    
     
\end{enumerate}
On the interval $[T_1,T_2]$, there are many functions that satisfy conditions 1-4, such as trigonometric functions 
\begin{align}\label{eq:cos}
\lambda_{2}(t)  &= \cos^2\left(\frac{\pi}{2} \cdot \frac{t -T_1}{T_2 -T_1}\right), & t \in [T_1, T_2],\\
\eta_{2}(t) &= \sin^2\left(\frac{\pi}{2} \cdot \frac{t -T_1}{T_2 -T_1}\right), & t \in [T_1, T_2].
\end{align}
Nevertheless, given that conditions 1-4 readily allow for the determination of a unique 
 cubic polynomial on the interval $[T_1,T_2]$, we are inclined to select the following cubic polynomial as our influence and adjoint  functions
 \begin{align}\label{eq:lambdaT1T2}
\lambda_{2}(t) &= 
2\left(\frac{t- T_1}{T_2-T_1}\right)^3 - 3\left(\frac{t- T_1}{T_2-T_1}\right)^2+1, & t \in [T_1, T_2], \\
\eta_{2}(t) &= 
-2\left(\frac{t- T_1}{T_2-T_1}\right)^3 + 3\left(\frac{t- T_1}{T_2-T_1}\right)^2, & t \in [T_1, T_2]. 
\end{align}
In order to improve the method's scalability, based on  \eqref{cond_yyz1} and \eqref{cond_yyz2}, we   broaden  the domain of definition of $\lambda_{2}(t)$ and $\eta_{2}(t)$ as follows
\begin{align}\label{eq:lambda}
\lambda_{2}(t) &= 
\begin{cases} 
2\left(\frac{t- T_1}{T_2-T_1}\right)^3 - 3\left(\frac{t- T_1}{T_2-T_1}\right)^2+1, & t \in [T_1, T_2], \\
0, & t \in (T_2, +\infty),
\end{cases}  \\
\eta_{2}(t) &= 
\begin{cases} 
-2\left(\frac{t- T_1}{T_2-T_1}\right)^3 + 3\left(\frac{t- T_1}{T_2-T_1}\right)^2, & t \in [T_1, T_2], \\
1 ,& t \in (T_2, +\infty).
\end{cases} 
\end{align}
{ In fact, the polynomial can be viewed as an approximation of any function that satisfies conditions 1-4.}  Figure \ref{fig:lambda-eta} shows the graphs of the influence function and its adjoint  function when trigonometric functions and polynomials are used, respectively.

If we further require that the second derivative be continuous at the interval node, then condition $\lambda^{''}_2(T_1)=0,\lambda^{''}_2(T_2)=0$ must be satisfied, and the corresponding quintic polynomial is given by 
$$\lambda_2(t)=-6\left(\frac{t- T_1}{T_2-T_1}\right)^5+15\left(\frac{t- T_1}{T_2-T_1}\right)^4-10\left(\frac{t- T_1}{T_2-T_1}\right)^3+1.$$


Sample training points in the spatio-temporal   domain $\Omega\times[T_1,T_2]$, optimize the loss function \eqref{loss_HDTime}, where $u_H$ is replaced by $u_{H_2}$, obtain the optimized value  
$\hat{u}_{\theta_2}$
  of $u_{\theta_2}$, thereby the PINN solution on the interval $[T_1,T_2]$ are obtained
\begin{equation}\label{eq:PINNS_2sub}
     \hat{u}_{H_2}(x,t) = \lambda_2(t)\cdot u_{\hat\theta_1}(x,t) + \eta_2(t)\cdot \hat{u}_{\theta_2}(x,t),\quad t\in [T_1,T_2].
\end{equation}

   \textbf{\textit{The hard constraint formulation \eqref{eq:HC_YYZ3} not only provides a reasonable initial data for the current interval by incorporating information from the previous interval, ensuring that the training process adheres to causality, but also strictly maintains the continuity and smoothness that   the exact solution of the equation has at time interval nodes. Furthermore, our method does not affect the PINN solution of the previous interval at all during the training of the current interval. While the method in \cite{guo2023pretraining} can also maintain the overall continuity and smoothness of the solution, it loses the solutions on the previous interval obtained through pre-training.}}

\subsubsection{Hard constraint learning strategy for partitioning the temporal domain  into multiple intervals}

We now aim to extend the method to accommodate three intervals instead of two. Given that  the temporal domain  $[T_0,T]$ is divided into three intervals, i.e.，
$$[T_0, T]=[T_0, T_1]\bigcup[T_1,T_2]\bigcup[T_2,T_3],\quad T_3=T,$$
and using the method from the previous subsection, we have successfully obtained the PINN solution \eqref{eq:PINNS_2sub} for the interval $[T_1,T_2]$. Next, we derive   the PINN solution for the third interval. By adopting the same hard constraint strategy employed for the second interval, we can establish the hard constraint format for the interval $[T_2,T_3]$ as follows
\begin{equation}\label{eq:HC_sub3}
     u_{H_3}(x,t) = \lambda_3(t)\cdot \hat{u}_{H_2}(x,t) + \eta_3(t)\cdot u_{\theta_3}(x,t),\quad t\in [T_2,T_3].
\end{equation}

 Noting \eqref{cond_yyz1}, when extending \eqref{eq:PINNS_2sub} to the interval $[T_2,T_3]$, we have 
\begin{equation}\label{eq:HC_T2}
\begin{aligned}
     \hat{u}_{H_2}(x,t) &= \lambda_2(t)\cdot u_{\hat\theta_1}(x,t) + \eta_2(t)\cdot \hat{u}_{\theta_2}(x,t)\\
     &= \hat{u}_{\theta_2}(x,t), \quad t\in[T_2,T_3].
     \end{aligned}
\end{equation}
 Thus, \eqref{eq:HC_sub3} is transformed into
\begin{equation}\label{eq:HC_sub31}
\begin{aligned}
    u_{H_3}(x,t) 
    = \lambda_3(t)\cdot  \hat{u}_{\theta_2}(x,t) + \eta_3(t)\cdot u_{\theta_3}(x,t), \quad t\in [T_2,T_3]. 
\end{aligned}  
\end{equation}

For $\lambda_3(t)$ and $\eta_3(t)$, we require them to satisfy conditions 1-4 in Section \ref{sec:2sub}, with a temporal shift where $T_1$
  is replaced by $T_2$
  and $T_2$
  by $T_3$.

  Sample training points in the spatio-temporal domain $\Omega\times[T_2,T_3]$, following the same training procedure as the second interval, obtain the optimized value  
$\hat{u}_{\theta_3}$
  of $u_{\theta_3}$, thereby obtaining the PINN solution on the interval $[T_2,T_3]$
\begin{equation}\label{eq:PINNS_3sub1}
     \hat{u}_{H_3}(x,t) = \lambda_3(t)\cdot  \hat{u}_{\theta_2}(x,t) + \eta_3(t)\cdot \hat{u}_{\theta_3}(x,t),\quad t\in [T_2,T_3].
\end{equation}
\eqref{cond_yyz2} ensures that the first-order derivatives of the PINN solutions \eqref{eq:PINNS_2sub} and \eqref{eq:PINNS_3sub1} over the two intervals are strictly equal at $T_2$
  with respect to the time variable $t$. 
  By extension, using the same method, we can obtain the PINN solution for the subsequent intervals.

 \textbf{\textit{ From the above analysis and \eqref{eq:PINNS_3sub1}, it is evident that for the third interval, only the training results  
$ \hat{u}_{\theta_2}(x,t)$ on the second interval are required, without needing the training results  
$u_{\hat\theta_1}(x,t)$ on the first interval. This makes our method easily extensible. When the temporal domain  needs to be extended, only the training results on the last interval are needed, making the algorithm concise and efficient.}}

\subsubsection{Adjustable influence  and adjoint  functions}

From the hard constraints   \eqref{eq:HC_YYZ3} and \eqref{eq:HC_sub31}, it is evident that their essence is to incorporate the PINN solution from the previous interval as a crucial component into the PINN solution of the current interval. This approach, however, leads to a problem: if the solution of the equation has a weak extrapolation ability in the time direction, 
that is, there is a significant difference in the form of the solution between two adjacent intervals, then this combination mechanism will make the training of the current interval particularly difficult. 
The reason is that \textbf{\textit{the influence function acts as a weight, determining the importance of the current interval in the training process. When the PINN solution of the previous and current intervals differ significantly, and the weight of the current interval is small, the resulting PINN solution for the current interval becomes very inaccurate.}}
To address this, we introduce parameters into the influence function and its adjoint function, making the influence interval adjustable. Below, taking the influence function $\lambda_{i}(t)$ and adjoint function $\eta_{i}(t)$ for the interval $[T_{i-1}, T_{i}],i\ge2$, where $i\geq 2$, as examples, we provide the computational formulas for modifying them into adjustable influence and adjoint functions.

Introducing the parameter $p_{i} \in(T_{i-1},T_{i}]$, the adjustable influence and adjoint functions are set as follows, with the adjustable trigonometric functions being  
\begin{align}
\lambda_{i}(t,p_{i}) &= 
\begin{cases}
\cos^2\left(\frac{\pi}{2} \cdot \frac{t -T_{i-1}}{p_{i} -T_{i-1}}\right), & t \in [T_{i-1}, p_{i}], \\
0, & t \in (p_{i}, +\infty),
\end{cases}   \\
\eta_{i}(t,p_{i}) &= 
\begin{cases}
\sin^2\left(\frac{\pi}{2} \cdot \frac{t -T_{i-1}}{p_{i} -T_{i-1}}\right), & t \in [T_{i-1}, p_{i}], \\
1 ,& t \in (p_{i}, +\infty).
\end{cases}
\end{align}
with the adjustable polynomial  functions being
\begin{align}\label{eq:lambda}
\lambda_{i}(t,p_{i}) &= 
\begin{cases} 
2\left(\frac{t- T_{i-1}}{p_{i}-T_{i-1}}\right)^3 - 3\left(\frac{t- T_{i-1}}{p_{i}-T_{i-1}}\right)^2+1, & t \in [T_{i-1}, p_{i}], \\
0, & t \in (p_{i}, +\infty),
\end{cases}  \\
\eta_{i}(t,p_{i}) &= 
\begin{cases} 
-2\left(\frac{t- T_{i-1}}{p_{i}-T_{i-1}}\right)^3 + 3\left(\frac{t- T_{i-1}}{p_{i}-T_{i-1}}\right)^2, & t \in [T_{i-1}, p_{i}], \\
1 ,& t \in (p_{i}, +\infty).
\end{cases} 
\end{align}

\begin{figure}[htbp]
    \centering
\includegraphics[width=0.9\textwidth]{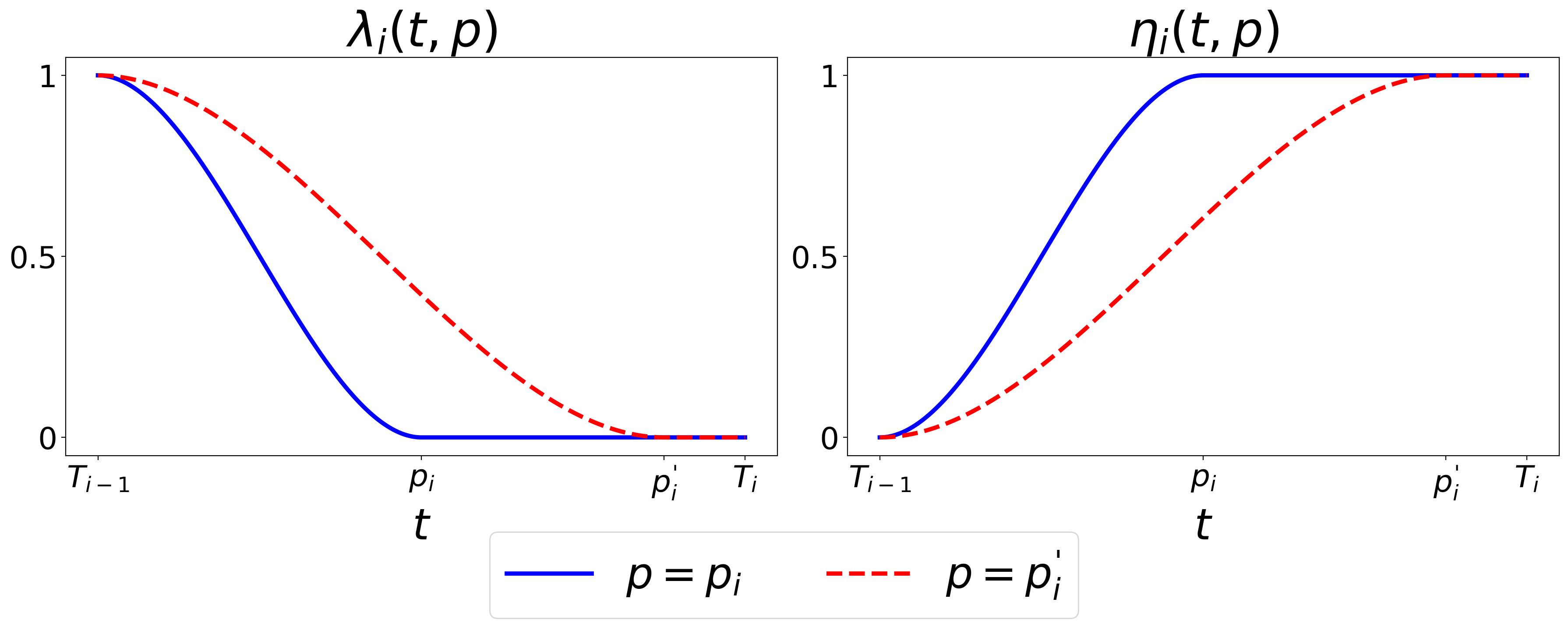}
\caption{Influence functions and adjoint functions with different $p_i$.}
\label{fig:lambda-eta-pq}
\end{figure}

Figure \ref{fig:lambda-eta-pq} presents the graphs of the influence function and the adjoint function for different values of $p_{i}$. The magnitude of $p_{i}$ is related to the property of the specific governing equation. If the solution of the equation has weak extrapolation ability in the time direction, its value should be close to $T_{i-1}$. This implies that the solution in the current interval is only influenced by the previous interval near $T_{i-1}$. 

Since it is generally impossible to determine in advance the extrapolation  ability of the equation's solution in the time direction, we need to treat this parameter as a trainable hyperparameter so that it can adapt to specific problems. 

\begin{remark}
In practical examples, we set the initial value of $p_{i}$ to the midpoint of the interval $[T_{i-1},T_{i}]$. Numerical experiments show  that for the problems with weak extrapolation, $p_{i}$ will move towards $T_{i-1}$
  during the training process. 
  \end{remark}
  
\textbf{\textit{ Integrating all the aforementioned techniques, we name this hard constraint sequential learning method with trainable influence functions as THC-PINNs (Trainable Hard Constraint  PINNs), and
the method with fixed parameter in the influence functions as FHC-PINNs (Fixed-parameter
Hard Constraint PINNs). 
}}
  
\section{Adaptive temporal domain partitioning algorithm }\label{sec3}

For sequential learning methods, choosing the appropriate interval length to partition the entire temporal domain  into a suitable number of intervals is a crucial task. If the intervals are too large, PINNs may fail to train effectively   on the first interval. Even if a usable predictive function can be  obtained, low accuracy will affect the training accuracy of subsequent intervals and may even lead to training failure. Conversely, if the intervals are too small, the total number of intervals increases, leading to many training sessions. This will not only significantly reduce computational efficiency but also result in severe cumulative errors, affecting the overall computational accuracy.

Although numerous researchers have been dedicated to developing efficient and high accuracy sequential deep learning methods for evolutionary equations,  {to our knowledge}, there has been no effective algorithm   for the reasonable   partitioning of the temporal domain. To address this issue, we  design a domain partitioning  algorithm based on a posterior error, providing an effective technical means for the rational partitioning of the overall temporal domain. \textbf{\textit{The core idea of this algorithm is that if the PINN solution trained on the interval $[0,T]$ is valid, it should yield consistent predictive results on the interval $[0,\frac{T}{2}]$ compared to the PINN solution trained on the interval $[0,\frac{T}{2}]$.}}
The specific algorithm   for determining  the time interval length is given below.
\begin{enumerate}
    \item 
\textbf{Preparation}
\begin{itemize}
    \item 
    Set the initial length of the interval to 
$T=T_{init}$, which can be any predicted value. It can be taken as the total duration or its half,
    \item 
     Set the threshold $\delta=5.0\times10^{-3}$ or $\delta=1.0\times10^{-3}$ for measuring the approximate error,
    \item
    Set the initial value of the parameter $\mathcal{D}_{last}$ as 
    $\mathcal{D}_{last}=1.0\times10^{15}$,
    \item 
    Obtain the PINN solution over the interval $[0,T]$ by using the standard PINN method, denote as 
   $u_\theta^{T}(x,t)$.
\end{itemize}

\item 
\textbf{Solve the equation by halving the interval length
 }\\
    Use the standard PINN method, obtain the PINN solution on the interval $[0,\frac{T}{2}]$, denoted as $u_\theta^{\frac{T}{2}}(x,t)$.

\item 
\textbf{Select collocation points}\\
Select $M$ collocation points in the corresponding spatio-temporal domain to form a set  
\begin{align*}
\tau_{test} = \left\{ \left( x_i, t_i \right) \left|_{i=1}^{M}\right  (x_i, t_i) \in \Omega \times \left[0,\frac{T}{2}\right] \right\}.
\end{align*}

\item
\textbf{Calculate the approximate $L_2$ relative error}\\
Calculate the posterior error using the collocation points with the following relative $L_2$ error formula
\begin{equation}\label{YYYYZ}
\mathcal{D} = \frac{\sqrt{\sum_{i=1}^{M} \left| u_\theta^{T}(x_{i}, t_{i}) - u_\theta^{\frac{T}{2}}(x_{i}, t_{i}) \right|^{2}}}{\sqrt{\sum_{i=1}^{M} \left| u_\theta^{\frac{T}{2}}(x_{i}, t_{i}) \right|^{2}}}.
\end{equation}

\item 
\textbf{Determine interval length}
\begin{itemize}
    \item 
     If $\mathcal{D} >\delta$ and
$\mathcal{D} \le \mathcal{D}_{last}$, then update the criteria. Let
\begin{align}
    &T = \frac{T}{2}, \\
    &u_{\theta}^{T}(x,t) = u_{\theta}^{\frac{T}{2}}(x,t), \\
    &\mathcal{D}_{\text{last}} = \mathcal{D},
\end{align}
Then proceed to step 2 to perform the next iteration.
\item 
    If $\mathcal{D}\le\delta$ or $\mathcal{D}>\mathcal{D}_{last}$, 
    this indicates that an approximately identical PINN solution has been obtained on the common region of the two intervals. The prediction result of $u_\theta^{\frac{T}{2}}(x,t)$ on the interval $[0,\frac{T}{2}]$ has tended towards convergence. The reasonable length of the interval is set to $\frac{T}{2}$, and the selection of interval length is concluded.

\end{itemize}
\end{enumerate}

In the aforementioned algorithm, the reason for setting the parameter $\mathcal{D}_{last}$  is as follows: when the interval length is relatively small, the computational accuracy has already tended to stabilize. Further reducing the interval length may not only fail to decrease the approximate $L_2$  error but could also result in a slight increase.

\section{Numerical experiments}\label{sec4}

In this section, we validate the effectiveness of the newly proposed hard constraint strategy by computing several typical evolutionary equations, including the one-dimensional convection equation, the Allen-Cahn equation, and the Korteweg–de Vries equation, which are widely used as benchmarks in studying the characteristics of PINNs \cite{RAISSI2019686,Mojgani2022LagrangianPA,Krishnapriyan2021CharacterizingPF,PENWARDEN2023112464,krishnapriyan2021characterizing,Wang2024RespectingCF,JUNG2024117036}.

We conduct the experiments using the PyTorch framework (version 2.5.1), with data types set to \texttt{float32} and the activation function set to \texttt{tanh}. The optimization process initially employs the Adam optimizer \cite{kingma2017adam} for 5000 iterations, followed by the L-BFGS optimization algorithm \cite{LLLBFGS} until convergence. For the parameters and stopping criteria of the L-BFGS optimization algorithm, we follows the settings from \cite{LLLBFGS}. The network architectures and hyperparameters used in the examples are detailed in Table \ref{tab:hyperparameters}.
\begin{table}[htbp]
\centering
\begin{tabular}{c|ccccc}
\hline
Equation & Architecture& Depth &  Width  & $w_i,w_b,w_r$ \\ \hline
Convection & Fourier feature NN& 4 & 40 & 1,1,1 \\ 
Allen-Cahn & Fourier
feature NN & 4 & 40 & 100,1,1\\ 
KdV & Fully connected NN & 3 & 50 & 1,1,1\\ 
Heat & Fully connected NN & 3 & 50 & 1,1,1\\ 
\hline
\end{tabular}
\caption{Network hyperparameter settings in numerical experiments.}
\label{tab:hyperparameters}
\end{table}

We evaluate the predictive accuracy of the method using the $L_2$
  relative error, $L_1$
  error, and $L_{\infty}$
  error, which are defined as follows.
  \begin{align}
&\left\|\epsilon\right\|_{2}= \frac{\sqrt{\sum_{i=1}^{N}\left|u_{\theta}(x_{i},t_{i})-u(x_{i},t_{i})\right|^{2}}}{\sqrt{\sum_{i=1}^{N}\left|u(x_{i},t_{i})\right|^{2}}}, \label{eq:l2_norm} \\
 &\|e\|_1 = \frac{1}{N} \sum_{i=1}^{N} \left| u_\theta(X_i, t_i) - u(X_i, t_i) \right|, \label{eq:l1_norm} \\
 &\|e\|_\infty = \max_{1 \leq i \leq N} \left| u_\theta(X_i, t_i) - u(X_i, t_i) \right|. \label{eq:linf_norm}
\end{align}
  Here, $u(x_{i},t_{i})$ represents the analytical or reference solution at the sample point $(x_{i},t_{i})$, while $u_{\theta}(x_{i},t_{i})$ denotes the PINN solution at that point. $N$ is the number of sample points. 
  
   Based on Remark \ref{remark1}, in all examples, for the first time interval $[0,T_1]$, training is carried out using a soft constraint method.

\subsection{Convection equation}\label{sec:Convection equation}
The convection equation is one of the fundamental equations in hydrodynamics and thermodynamics, and it has widespread applications in fields such as meteorology, oceanography, environmental science, and engineering. These equations are used to describe the process of physical quantities (such as mass, momentum, energy, etc.) being transported by the movement of a fluid.

We consider the following one-dimensional convection equation
\begin{equation}\label{eq:Convection equation}
\begin{cases}
u_t-\beta u_x=0, &(x,t) \in (0,2\pi) \times  (0,T],\\
u(x,0)= \sin{ x},& x\in   [0,2\pi],   \\
u(0,t)=u(2\pi,t),& t\in  (0,T],
\end{cases}
\end{equation}
where $\beta$ is the convection coefficient, and here we set $\beta=40$. The exact solution of \eqref{eq:Convection equation} is given by  
\begin{equation*}\label{eq:Convection Exact}
u(x,t)=\sin{(x-\beta t)}.
\end{equation*}

First, we examine the relationship between   $p_i$
   in the adjustable influence function $\lambda(t,p_i)$ determined by \eqref{eq:lambda} and the accuracy. Let $T=2.0$, which means we aim to obtain the PINN solution over the temporal domain 
 $[0,2]$. The  domain  is divided into two intervals: $$[0,2]=[0,1]\cup(1,2].$$
  We employ a sequential learning approach to train and obtain the corresponding PINN solutions on each interval.
For the interval $[0,1]$, we use the standard PINN method to obtain its predictive function. For the second interval $[1,2]$, we solve it using the hard constraint method, where the adjustable parameter $p_i$
  ranges within $$p_i\in(1,2].$$
  
\begin{figure}[htbp]\label{AE_fig1}
\centering
\includegraphics[width = 1.0\textwidth]{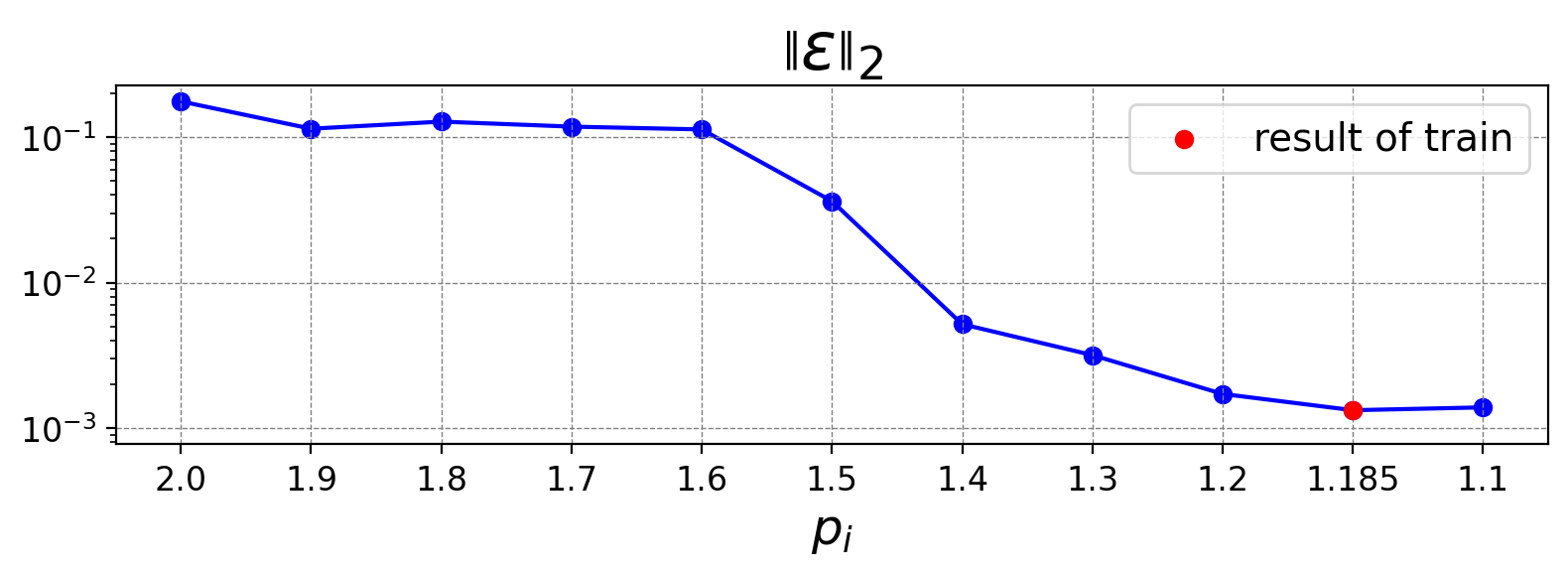}
\caption{The  $L_2$ relative
errors with different $p_i$ for the convection equation.}
\end{figure}

Table \ref{AE-table} and Figure \ref{AE_fig1}  present the computational accuracy over the entire temporal domain  $[0,2]$ obtained with different values of the adjustable parameter  $p_i$. It can be observed that, for this example, the closer the position of  $p_i$
  is to the left endpoint of the interval $(1,2]$, the higher the accuracy of the obtained PINN solution.
\textbf{\textit{The last row in Table \ref{AE-table} shows the PINN solution obtained by training $p_i$
  as a hyperparameter together with the network parameters, which achieves the optimal accuracy.}}
 \begin{table}[htbp]
    \centering
    \begin{tabular}{c|ccc}
        \hline
        $p_i$ &  $\left\|\epsilon\right\|_{2}$  & $\|e\|_1 $ & $\|e\|_\infty$ \\ 
        \hline
2.0 & $1.7603 \times 10^{-1}$ & $7.2193 \times 10^{-2}$ & $3.7676 \times 10^{-1}$ \\
1.9 & $1.1424 \times 10^{-1}$ & $4.7571 \times 10^{-2}$ & $2.3697 \times 10^{-1}$ \\
1.8 & $1.2781 \times 10^{-1}$ & $5.3010 \times 10^{-2}$ & $2.5071 \times 10^{-1}$ \\
1.7 & $1.1808 \times 10^{-1}$ & $4.8673 \times 10^{-2}$ & $2.2814 \times 10^{-1}$ \\
1.6 & $1.1295 \times 10^{-1}$ & $4.6196 \times 10^{-2}$ & $2.1811 \times 10^{-1}$ \\
1.5 & $3.6290 \times 10^{-2}$ & $1.4753 \times 10^{-2}$ & $8.5906 \times 10^{-2}$ \\
1.4 & $5.1759 \times 10^{-3}$ & $2.3466 \times 10^{-3}$ & $1.1470 \times 10^{-2}$ \\
1.3 & $3.1714 \times 10^{-3}$ & $1.4725 \times 10^{-3}$ & $7.4459 \times 10^{-3}$ \\
1.2 & $1.7148 \times 10^{-3}$ & $8.5310 \times 10^{-4}$ & $4.1729 \times 10^{-3}$ \\
1.1 & $1.3865 \times 10^{-3}$ & $7.3174 \times 10^{-4}$ & $2.9633 \times 10^{-3}$ \\
   \hline
 {\textbf{1.185(Training)}} & { $\mathbf{1.3258 \times 10^{-3}}$}& {$\mathbf{ 7.0547 \times 10^{-4}}$} & {$\mathbf{2.8410 \times 10^{-3}}$}  \\
        \hline
    \end{tabular}
    \caption{
    The computational errors with different $p_i$ for the convection equation $(T=2)$.}
    \label{AE-table}
\end{table}

\begin{figure}[htbp]
\centering
\includegraphics[width=1.0\textwidth]{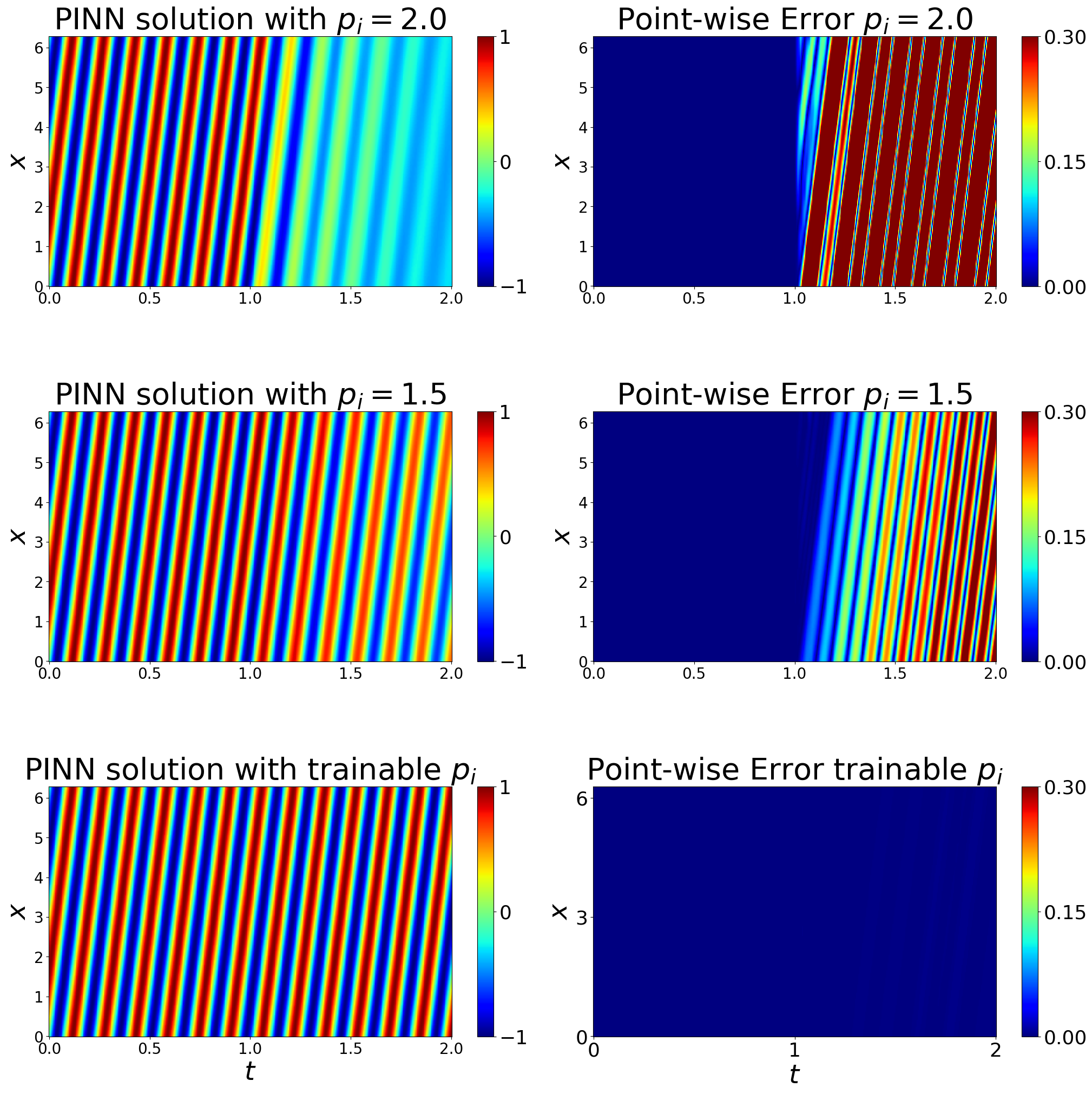}
\caption{
Comparison of the PINN solutions obtained with different $p_i$ and the exact solution $(T=2)$.}
\label{AE_fig2}
\end{figure}

Figure \ref{AE_fig2} presents a comparison between the PINN solutions obtained with $p_i=2$, $p_i=1.5$ and  {trainable $p_i$}  and the exact solution. It is evident that the PINN solution with $p_i=2$ significantly deviates from the exact solution, rendering it an invalid prediction. In contrast, the PINN solution obtained through training $p_i$
  closely approximates the exact solution. For specific problems, it is challenging to manually set an appropriate value for $p_i$. Therefore, training it as a trainable parameter is not only necessary but also feasible.

\begin{figure}[htbp]
\centering
\includegraphics[scale = 0.7]{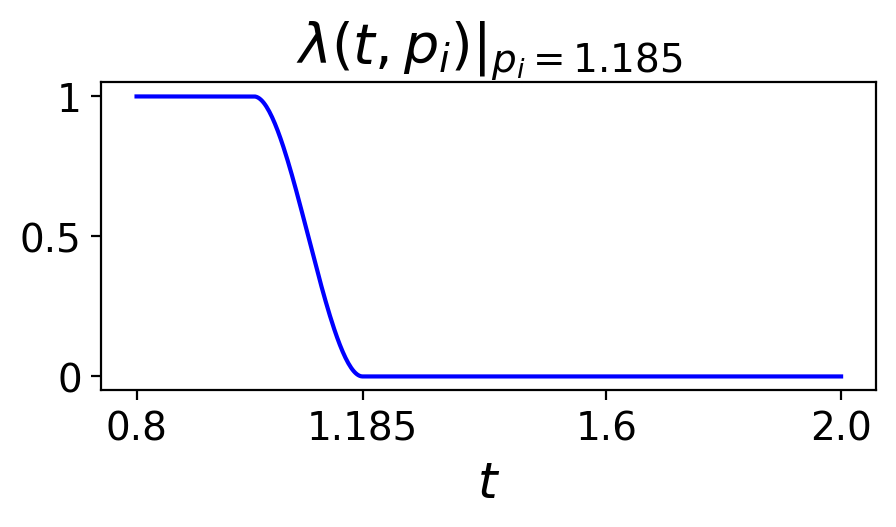}
\caption{The influence function $\lambda(t,p_i)|_{p_i=1.185}$.}
\label{AE_fig3}
\end{figure}

 Figure \ref{AE_fig3} displays the graph of the influence function $\lambda(t,p_i)|_{p_i=1.185}$
  obtained through training. It can be observed that within the interval [1,2], it rapidly decreases from $1$ to $0$. This indicates that the PINN solution for the interval $[0,1]$ is not suitable for representing the solution in the interval $[1,2]$, meaning that the PINN solution of \eqref{eq:Convection equation} has very weak extrapolation ability. \textbf{\textit{Actually， the convection equation is a hyperbolic type equation and does not have a smoothing effect, {which}  is also the reason for its weak extrapolation ability.}}

To investigate the capability of our method in handling large temporal domain, we set $T=5$. Table \ref{AE-multgroup} lists the computational accuracy of the PINN solution over the entire temporal domain  $[0,5]$ using different number of intervals. It can be seen that when the number of intervals exceeds $6$, the computational accuracy tends to stabilize.  {It should be emphasized that these results are obtained by training $p_i$
  as a trainable parameter. If $p_i$
  is fixed at the right endpoint of the interval, the training will completely fail when the number of intervals is large. }
\begin{table}[htbp]
    \centering
    \begin{tabular}{c|ccc}
        \hline
        Number of intervals &  $\left\|\epsilon\right\|_{2}$  & $\|e\|_1 $ & $\|e\|_\infty$ \\ 
        \hline
1（Standard PINNs） & $1.0084 \times 10^{0}$ & $6.2818 \times 10^{-1}$ & $1.1525 \times 10^{0}$ \\
2 & $3.2501 \times 10^{-1}$ & $1.8549 \times 10^{-1}$ & $4.3035 \times 10^{-1}$ \\
3 & $9.0707 \times 10^{-3}$ & $4.8846 \times 10^{-3}$ & $1.4975 \times 10^{-2}$ \\
4 & $7.6671 \times 10^{-3}$ & $3.9253 \times 10^{-3}$ & $1.5699
         \times 10^{-2}$ \\
5 & $7.8902 \times 10^{-3}$ & $4.1556 \times 10^{-3}$ & $1.5276 \times 10^{-2}$ \\
6 & $6.0261 \times 10^{-3}$ & $3.1697 \times 10^{-3}$ & $1.1848 \times 10^{-2}$ \\
7 & $5.5845 \times 10^{-3}$ & $2.9423\times 10^{-3}$ & $1.0881 \times 10^{-2}$ \\
8 & $6.6414 \times 10^{-3}$ & $3.2404 \times 10^{-3}$ & $1.5897 \times 10^{-2}$ \\
9 & $6.2175 \times 10^{-3}$ & $3.1028 \times 10^{-3}$ & $1.5483 \times 10^{-2}$ \\
10 & $6.6994 \times 10^{-3}$ & $3.3523 \times 10^{-3}$ & $1.5185 \times 10^{-2}$ \\
        \hline
    \end{tabular}
    \caption{Solving the convection equation using THC-PINNs with different number of intervals $(T=5)$.
    }
    \label{AE-multgroup}
\end{table}
 
  Table \ref{adpTTTT} provides the approximate $L_2$
  relative error $\mathcal{D}$ calculated according to  \eqref{YYYYZ}. If the threshold in the algorithm is set to $\delta=1.0\times10^{-2}$,   $4$ intervals should be used for  the calculation of the overall domain. If $\delta=1.0\times10^{-3}$, then $8$ intervals should be used for the calculation of the overall domain. Combining the results in Table \ref{AE-multgroup}, it can be seen that this adaptive time domain partitioning strategy is feasible.
\begin{table}[htbp]
    \centering
    \begin{tabular}{ccccc}
        \hline
  $T$ & $\frac{T}{2}$ & Initial interval & $\mathcal{D}$ &Number of intervals\\ 
        \hline
5.000 & 2.5000& $[0,2.5000]$ &$9.9493 \times 10^{-1} $&2\\
2.500&1.2500 & $[0,1.2500]$ & $1.5895 \times 10^{-3} $&4 \\
1.250&0.6250& $[0,0.6250]$& $8.0895 \times 10^{-4}$&8 \\
0.625&0.3125 & $[0,0.3125]$ & $3.9079 \times 10^{-4} $ &16\\
        \hline
    \end{tabular}
    \caption{Adaptive time domain partitioning for the convection equation.}
    \label{adpTTTT}
\end{table}

\begin{figure}[htbp]
\centering
\includegraphics[width=1.0\textwidth]{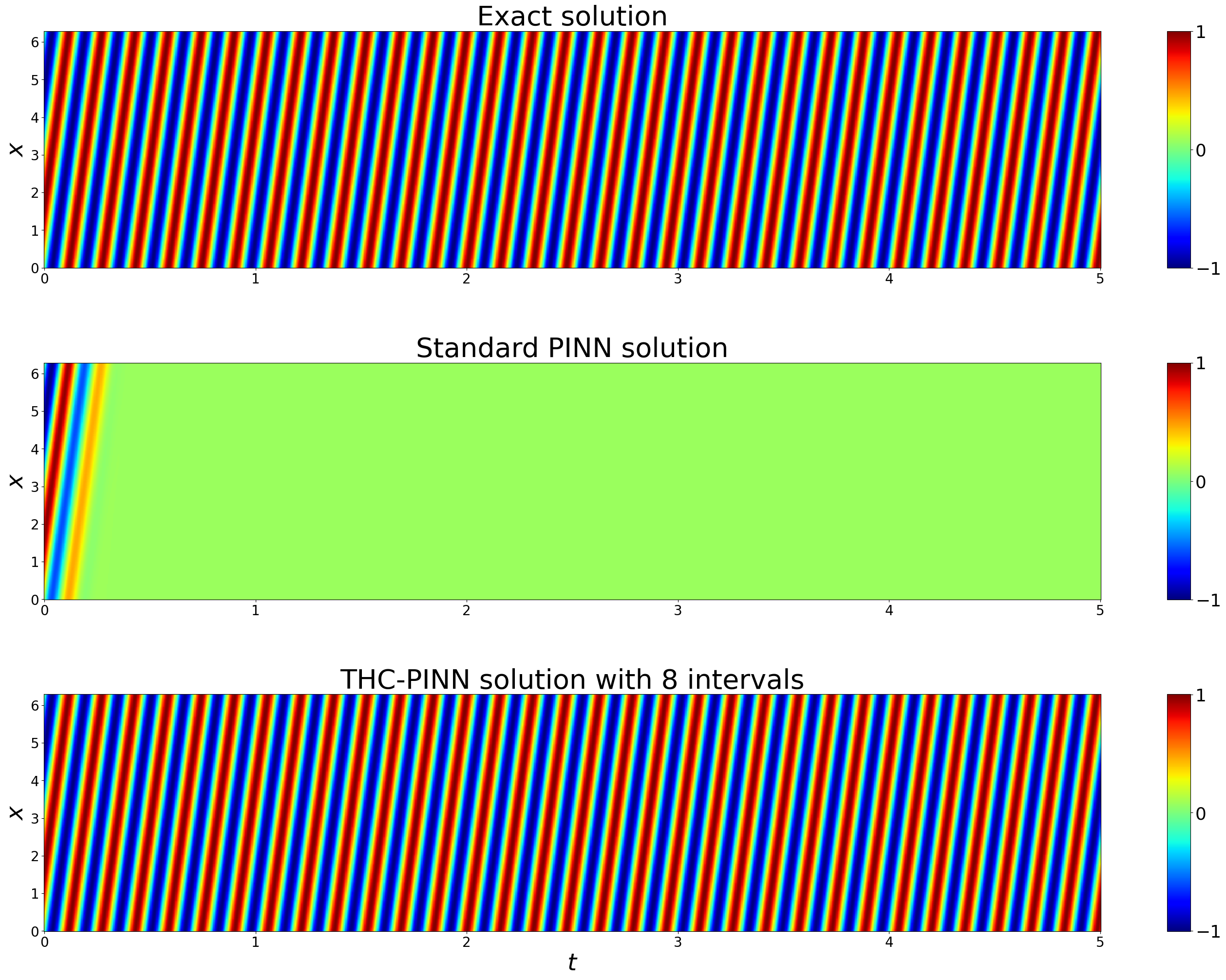}
\caption{Comparison of different PINN solutions and the exact solution for the convection equation $(T=5)$.}
\label{conv40_c1}
\end{figure}

\begin{figure}[htbp]
\centering
\includegraphics[width=1\textwidth]{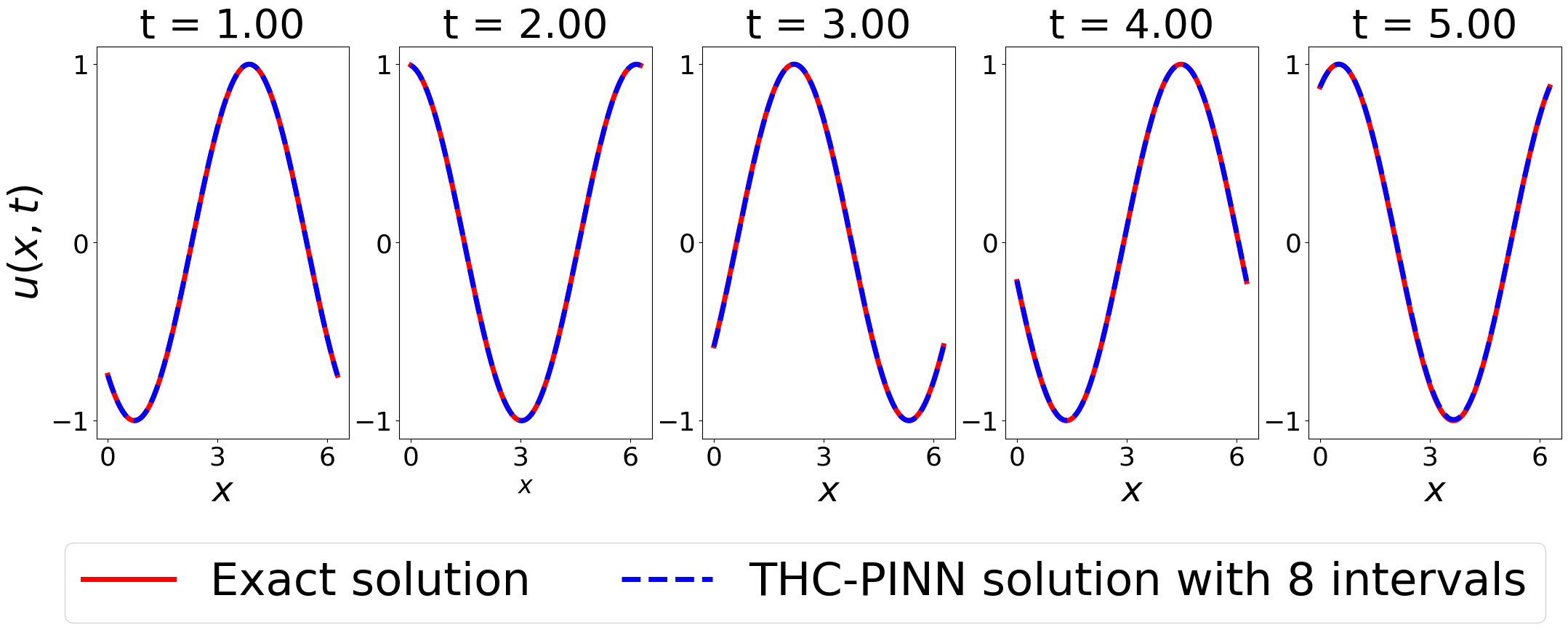}
\caption{Comparison of the THC-PINN solution (with $8$ intervals) and the exact solution of the convection equation at different times.}
\label{conv40_c2}
\end{figure}

Figures \ref{conv40_c1} and \ref{conv40_c2} show the comparison between the THC-PINN solution and the exact solution. It is evident that we have obtained an effective PINN solution over the long temporal domain  $[0,5]$. Notably, the THC-PINN solution here uses $8$ intervals, indicating that our method does not exhibit significant error accumulation when the number of intervals is large.

\subsection{Allen-Cahn 
 equation}\label{sec:Allen-Cahn equation}

 The Allen-Cahn (AC) equation is a class of important partial differential equations that find extensive applications in materials science, physical chemistry, and image processing, among other fields. It is used to describe the evolution of phase transition interfaces and changes in material properties, which is of significant importance for the preparation and performance improvement of materials. In PINN benchmark tests, a commonly used form is   shown by
\begin{equation}\label{eq:AC equation}
\begin{cases}
u_t-0.0001u_{xx}+5u^3-5u=0, &(x,t) \in (-1,1) \times  (0,T],\\
u(x,0)=x^2 \cos{\pi x},& x\in  [-1,1],   \\
u(-1,t)=u(1,t), u_x(-1,t)=u_x(1,t), & t\in  (0,T].\\
\end{cases}
\end{equation}

Similar to the previous example, we first examine the relationship between t $p_i$
 in the adjustable influence function $\lambda(t,p_i)$ determined by \eqref{eq:lambda}  and the accuracy. Here, we set $T=0.5$. We divide the temporal 
 domain into two intervals and then employ a sequential learning approach to train and obtain the corresponding PINN solutions on each interval.
  
For the first interval $[0,0.25]$, we use the standard PINN method to obtain its predictive function. For the second interval $[0.25,0.5]$, we solve it using the hard constraint method, where the adjustable parameter $p_i$
  ranges within $$p_i\in(0.25,0.5].$$

\begin{figure}[htbp]
\centering
\includegraphics[width = 1.0\textwidth]{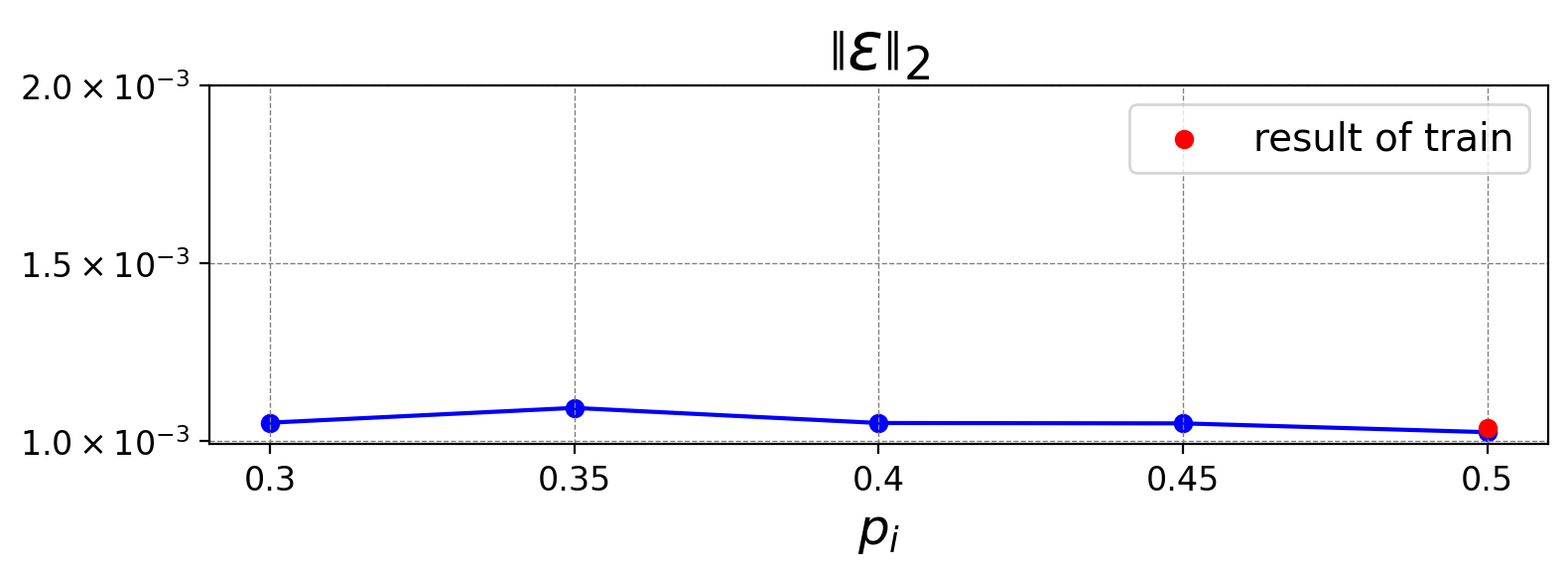}
\caption{The $L_2$ relative errors with different $p_i$ for the  AC equation.}
\label{AC_Ac_PI}
\end{figure}

 Table \ref{Table_AC-PI} and Figure \ref{AC_Ac_PI} present the computational accuracy over the entire temporal domain  $[0,0.5]$ obtained with different values of the adjustable parameter $p_i$.  {Contrary to the previous example, the position of $p_i$
  has almost no impact on the accuracy of the PINN solution for this case. The last row in  Table \ref{Table_AC-PI} shows the results obtained by training $p_i$
  as a hyperparameter together with the network parameters, noting that  
$p_i$
  is trained to the right endpoint of the interval.}
\begin{table}[htbp]
    \centering
    \begin{tabular}{c|ccc}
        \hline
      $p_i$ &  $\left\|\epsilon\right\|_{2}$  & $\|e\|_1 $ & $\|e\|_\infty$ \\ 
        \hline
0.30 & $1.0509 \times 10^{-3}$ & $3.6181 \times 10^{-4}$ & $5.3887 \times 10^{-3}$ \\
0.35 & $1.0924 \times 10^{-3}$ & $3.7311 \times 10^{-4}$ & $5.3526 \times 10^{-3}$ \\
0.40 & $1.0500 \times 10^{-3}$ & $3.6229 \times 10^{-4}$ & $5.3230 \times 10^{-3}$ \\
0.45 & $1.0489 \times 10^{-3}$ & $3.5777 \times 10^{-4}$ & $5.3325 \times 10^{-3}$ \\
\hline
 {\bf 0.50(Training)} &   ${\bf 1.0372 \times 10^{-3}}$ &   ${\bf 3.5572 \times 10^{-4}}$ &  ${\bf 5.2209 \times 10^{-3}}$ \\
        \hline
    \end{tabular}
    \caption{The computational errors with different $p_i$ for the AC equation  $(T=0.5)$.}
    \label{Table_AC-PI}
\end{table}

\begin{figure}[htbp]
\centering
\includegraphics[scale = 0.7]{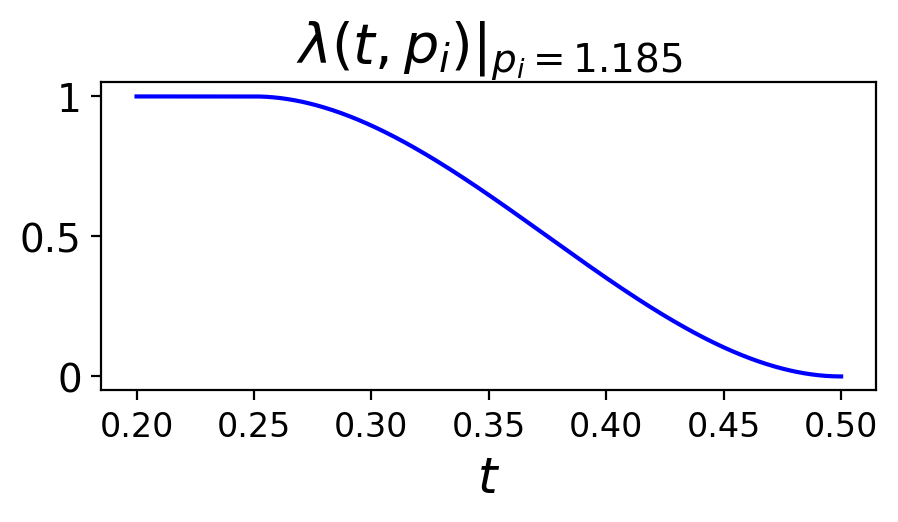}
\caption{The influence function $\lambda(t,p_i)|_{p_i=0.5}$.}
\label{AC_train}
\end{figure}

Figure \ref{AC_train} displays  the influence function $\lambda(t,p_i)|_{p_i=0.5}$
  obtained through training. It can be seen that within the interval $[0.25,0.5]$, it gently decreases from $1$ to $0$. This indicates that for this example, the PINN solution for the interval $[0,0.25]$ can well represent the solution for the interval $[0.25,0.5]$, meaning that the PINN solution of \eqref{eq:AC equation} has strong extrapolation ability. \textbf{\textit{The  AC equation, being a parabolic type equation with good smoothness, is the main reason for its strong extrapolation ability. For equations with strong extrapolation ability, the accuracy of the PINN solution is not sensitive to the position of $p_i$, and high computational accuracy can be achieved.}}

We extend the temporal domain  by setting $T=1$. Table \ref{AC-multgroup} lists the computational accuracy of the PINN solution over the entire temporal domain  $[0,1]$ obtained using different numbers of intervals with the training of $p_i$. {For this example, if $p_i$
  is fixed at the right endpoint of the interval, the computational accuracy is basically the same as when $p_i$
  is involved in the training,} as shown in Table \ref{AC_fix_multgroup}.

\begin{table}[htbp]
    \centering
    \begin{tabular}{c|ccc}
        \hline
        Number of intervals &  $\left\|\epsilon\right\|_{2}$  & $\|e\|_1 $ & $\|e\|_\infty$ \\ 
        \hline
1(Standard PINNs) & $6.3621 \times 10^{-2}$ & $1.3272 \times 10^{-2}$ & $5.7280 \times 10^{-1}$ \\
2 & $3.5347 \times 10^{-3}$ & $9.4887 \times 10^{-4}$ & $3.6919 \times 10^{-2}$ \\
3 & $3.5269 \times 10^{-3}$ & $8.7828 \times 10^{-4}$ & $4.1569 \times 10^{-2}$ \\
4 & $3.6812 \times 10^{-3}$ & $1.0197 \times 10^{-3}$ & $4.2303 \times 10^{-2}$ \\
5 & $3.9502 \times 10^{-3}$ & $1.0007 \times 10^{-3}$ & $4.5792 \times 10^{-2}$ \\
        \hline
    \end{tabular}
    \caption{Solving the AC equation using THC-PINNs with different number of intervals $(T=1)$.}
    \label{AC-multgroup}
\end{table}

\begin{table}[htbp]
    \centering
    \begin{tabular}{c|ccc}
        \hline
        Number of intervals &  $\left\|\epsilon\right\|_{2}$  & $\|e\|_1 $ & $\|e\|_\infty$ \\ 
        \hline
1(Standard PINNs) & $6.3622 \times 10^{-2}$ & $1.3272 \times 10^{-2}$ & $5.7280 \times 10^{-1}$ \\
2 & $4.6311 \times 10^{-3}$ & $1.1498 \times 10^{-3}$ & $5.2558 \times 10^{-2}$ \\
3 & $3.7752 \times 10^{-3}$ & $9.0799 \times 10^{-4}$ & $4.6615 \times 10^{-2}$ \\
4 & $3.6201 \times 10^{-3}$ & $8.5046 \times 10^{-4}$ & $4.5536 \times 10^{-2}$ \\
5 & $3.2316 \times 10^{-3}$ & $8.9031 \times 10^{-4}$ & $3.3212 \times 10^{-2}$ \\
        \hline
    \end{tabular}
    \caption{
    Solving the AC equation with different number of intervals where $p_i$ is
  fixed at the right endpoint  $(T=1)$.}
    \label{AC_fix_multgroup}
\end{table}

Table \ref{AC_adaTTT} provides the approximate $L_2$
  relative error $\mathcal{D}$ used for adaptive partitioning. If the threshold in the algorithm is set to $\delta=1.0\times10^{-2}$,  $4$ intervals should be used for the overall calculation. If $\delta=1.0\times10^{-3}$
 , then $8$ intervals should be used for the overall calculation.
\begin{table}[htbp]
    \centering
    \begin{tabular}{ccccc}
        \hline
  $T$ & $\frac{T}{2}$ &  Initial interval & $\mathcal{D}$ &Number of intervals\\ 
        \hline
1.0000 & 0.5000& $[0,0.5000]$ &$1.8612 \times 10^{-2}$  &2\\
0.5000&0.2500 & $[0,0.2500]$ & $1.2713 \times 10^{-3} $&4 \\
0.2500&0.1250& $[0,0.1250]$& $6.4375 \times 10^{-4}$&8 \\
0.1250&0.0625 & $[0,0.0625]$ & $7.0378 \times 10^{-4}$ &16\\
        \hline
    \end{tabular}
    \caption{Adaptive time  domain partitioning for the AC equation.}
    \label{AC_adaTTT}
\end{table}

\begin{figure}[htbp]
\centering
\includegraphics[width=1\textwidth]{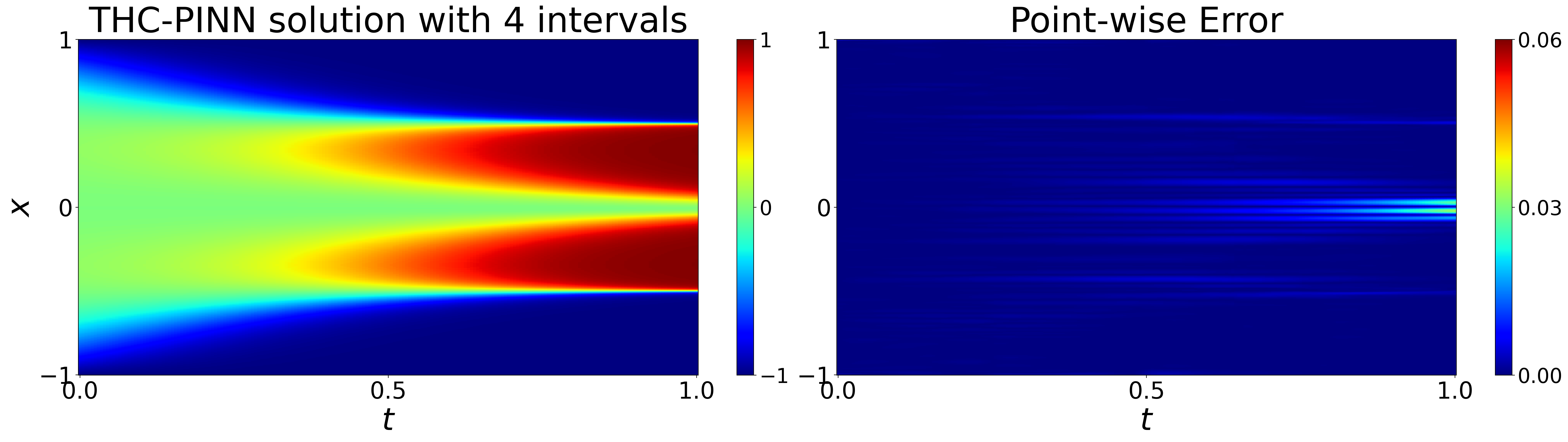}
\caption{
Comparison of THC-PINN solution and the reference solution for the AC equation
$(T=1)$.}
\label{AC_c1}
\end{figure}

\begin{figure}[htbp]
\centering
\includegraphics[width=1\textwidth]{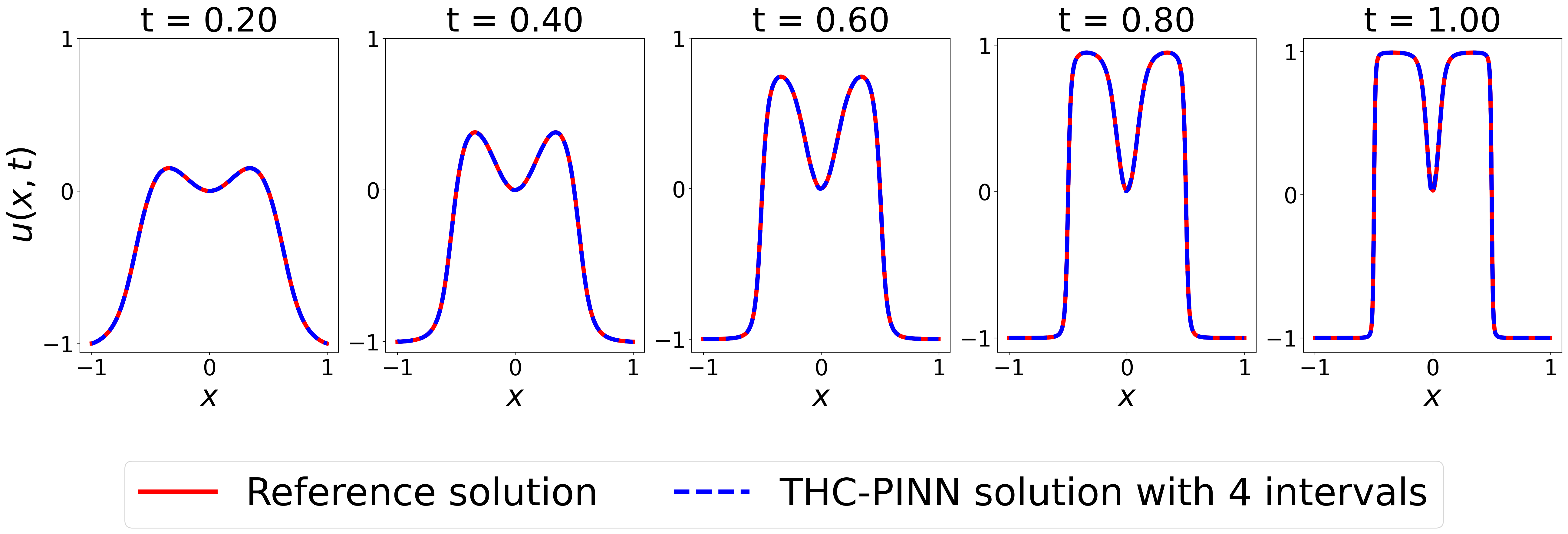}
\caption{
Comparison of the THC-PINN solution (using 4 intervals) and the reference solution of the
AC equation at different times.}
\label{AC_c2}
\end{figure}

Figures \ref{AC_c1} and \ref{AC_c2} show a comparison between the THC-PINN solution and the reference solution. We  obtain effective predictions on the temporal domain  $[0,1]$. Here, the THC-PINN solution uses $4$  intervals.

\subsection{Korteweg-de Vries 
 equation}\label{sec:KdV equation}

 The Korteweg-de Vries (KdV) equation is a significant nonlinear partial differential equation originally used to describe the propagation of long waves in shallow water, and it can well exhibit the nonlinear and dispersive characteristics of waves. This equation is widely applied in hydrodynamics, plasma physics, nonlinear optics, and other fields, holding significant importance for the study of nonlinear wave phenomena. 
 
 The KdV equation is a commonly used benchmark problem in the literature on PINNs, and a special form of this equation is as follows
\begin{equation}\label{eq:KdV equation}
\begin{cases}
u_t+uu_{x}+0.0025u_{xxx}=0, &(x,t) \in (-1,1) \times  (0,1],\\
u(x,0)=\cos{\pi x},& x\in  [-1,1],   \\
u(-1,t)=u(1,t), u_x(-1,t)=u_x(1,t), & t\in  (0,1].\\
\end{cases}
\end{equation}

Table \ref{AC_adaTTTTTT} lists the approximate $L_2$ relative error $\mathcal{D}$ used for adaptive partitioning. Based on this data, we solve the KdV equation using $4$ intervals. 
\begin{table}[htbp]
    \centering
    \begin{tabular}{ccccc}
        \hline
  $T$ & $\frac{T}{2}$ &  Initial interval  & $\mathcal{D}$ &Number of intervals\\ 
        \hline
1.0000 & 0.5000& $[0,0.5000]$ &$3.3025 \times 10^{-2}$  &2\\
0.5000&0.2500 & $[0,0.2500]$ & $5.4692 \times 10^{-3} $&4 \\
0.2500&0.1250& $[0,0.1250]$& $2.2238  \times 10^{-3}$&8 \\
0.1250&0.0625 & $[0,0.0625]$ & $1.3416 \times 10^{-3}$ &16\\
        \hline
    \end{tabular}
    \caption{Adaptive time interval partitioning for the KdV equation.}
    \label{AC_adaTTTTTT}
\end{table}

\begin{figure}[htbp]
\centering
\includegraphics[width=1\textwidth]{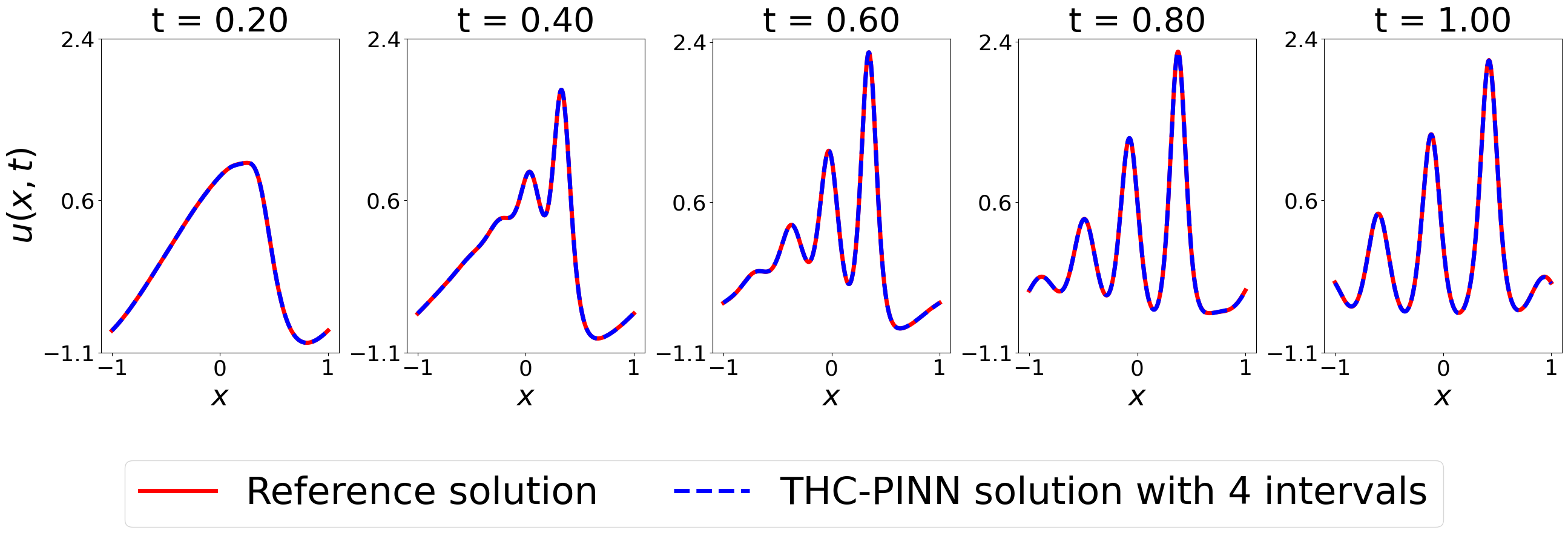}
\caption{
Comparison of the THC-PINN solution (using 4 intervals) and the reference solution of the
KdV equation at different times.}
\label{KdV_c2}
\end{figure}

\begin{figure}[htbp]
\centering
\includegraphics[width=1\textwidth]{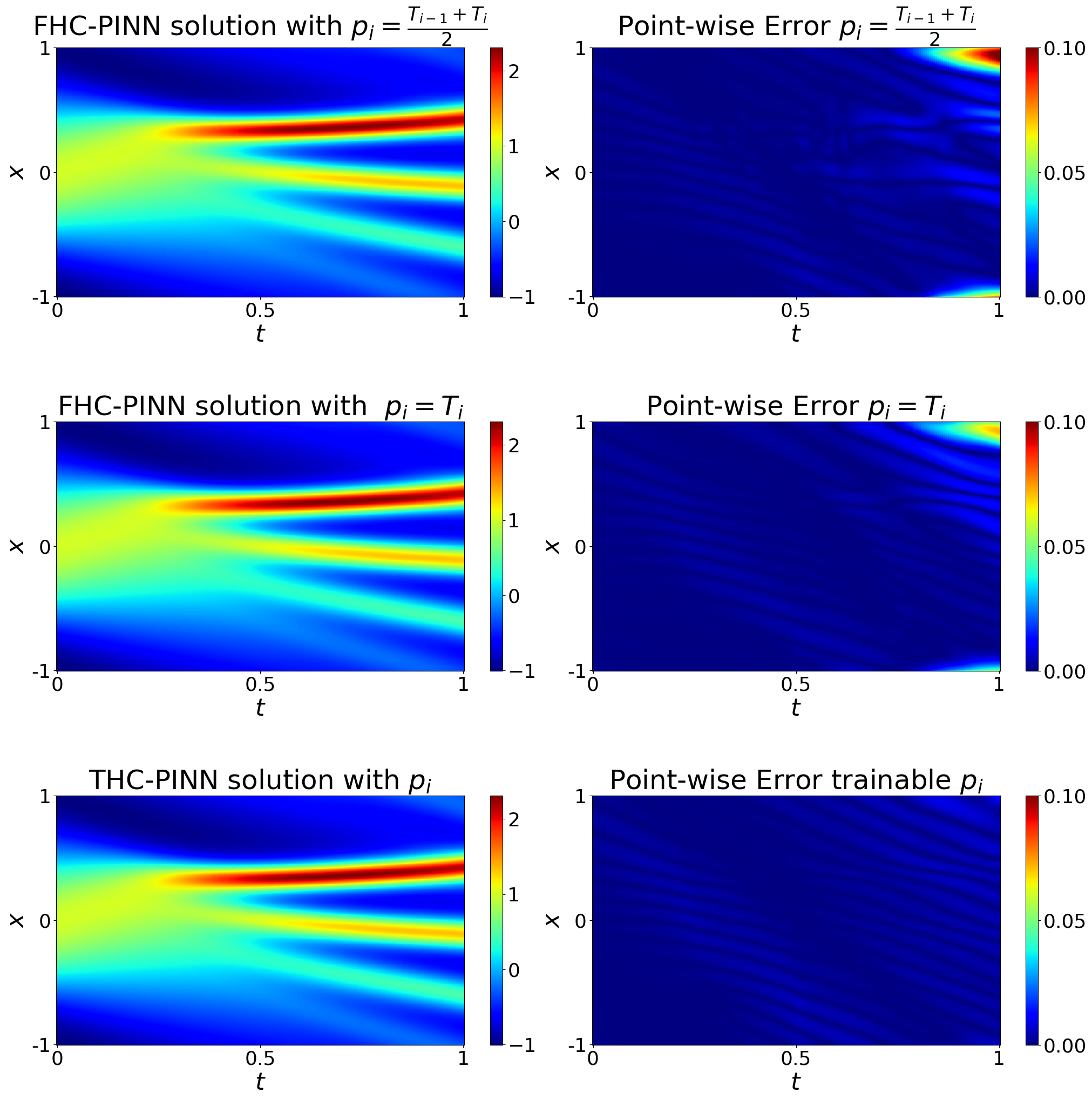}
\caption{
Comparison of the PINN solutions  with different $p_i$ and the reference solution $(T=1)$.}
\label{KdV_c1}
\end{figure}

Figure \ref{KdV_c2} shows the comparison between the THC-PINN solution and the reference solution. The predictive solution obtained by our method closely approximates the reference solution. Figure \ref{KdV_c1} and Table \ref{Table_KdV-PI} list the training results obtained with $p_i$ set at different positions using 4 intervals, which implies that THC-PINNs outperform FHC-PINNs.

\begin{table}[htbp]
    \centering
    \begin{tabular}{c|ccc}
        \hline
      $p_i$ &  $\left\|\epsilon\right\|_{2}$  & $\|e\|_1 $ & $\|e\|_\infty$ \\ 
    
\hline
 $ \frac{T_{i-1}+T_{i}}{2}$  & $1.0602 \times 10^{-2}$ & $2.9438 \times 10^{-3}$ & $7.1882 \times 10^{-2}$ \\
$T_{i}$& $8.5167 \times 10^{-3}$ & $2.4790 \times 10^{-3}$ & $5.4590 \times 10^{-2}$ \\
 {\textbf{Training}} &  {$\bf{5.6210 \times 10^{-3}}$} &  {$\bf{2.2025 \times 10^{-3}}$} &  {$\bf{2.9257 \times 10^{-2}}$} \\
        \hline
    \end{tabular}
    \caption{
    The computational errors with different  $p_i$ for the KdV equation 
 {$(T=1)$}.  } 
    \label{Table_KdV-PI}
\end{table}

\textbf{\textit{For the KdV equation, since it is neither a hyperbolic equation nor a parabolic equation, an interesting phenomenon occurs when training the parameter $p_i$ in the influence function: $p_i$ does not move towards the left endpoint of the interval like  the wave equation, nor does it move towards the right endpoint like  the AC equation, and the direction of movement is uncertain.}} In addition, 
  we also find that for this equation, when the number of intervals is large, there is a  accumulation of errors, and how to solve this issue is the work we need to study next.

\subsection{A three-dimensional heat conduction problem with large temporal gradients}

In materials science \cite{Vikhrenko_2011}, it is often necessary to study the thermal stability and durability of materials when the temperature changes dramatically in a short period of time.   Here we test the effectiveness of our method for the problem with large temporal gradients  by
\begin{equation*}\label{heat_equation}
\begin{cases}
u_t = \Delta u+f(\bm x,t), & \bm x \in \Omega, t \in (0,1], \\
u(\bm x, t)=g_0(\bm x), & t=0,\\
u(\bm x,t) =g_D(\bm x ,t), & \bm x\in\partial\Omega,
\end{cases}
\end{equation*}
here $\Omega=(-1,1)^3$, $g_D$, $g_0$ and $f$ are derived by the 
 exact solution 
\begin{equation*}
u(x,y,z,t) = (1.0 + 0.1  x   y   z)  e^{\frac{1}{(t-1)^2+\varepsilon}},
\end{equation*}
where $\varepsilon$ is a positive constant.
It can be seen that the variation of solutions in the spatiotemporal domain mainly depends on the size of $\varepsilon$ in the time term. If $\varepsilon$ is large, the variation of $u$ in the entire computational domain will be smooth; If $\varepsilon$ is very small, the value of $u$ will sharply change around $t=1$, reflecting the multi-scale characteristics in the time direction.


 Here we employ the proposed method THC-PINN 
 to solve this problem with $\varepsilon = 0.15$.  
 We adopt the adaptive temporal domain partitioning algorithm  in Section \ref{sec3} to   divide  temporal domain. Table \ref{Heat_adaTTTTTT} lists the approximate $L_2$ relative error $\mathcal{D}$ used for adaptive partitioning. Based on this data, we solve the three-dimensional heat conduction equation using $4$ intervals.   Notably,  the same as the AC equation,  the heat conduction equation also belongs to the parabolic type, and the solution has strong extrapolation ability.  $p_i$ is trained to the right endpoint of the intervals.



\begin{table}[htbp]
    \centering
    \begin{tabular}{ccccc}
        \hline
  $T$ & $\frac{T}{2}$ &  Initial interval  & $\mathcal{D}$ &Number of intervals\\ 
        \hline
       1.0000 & 0.5000 & $[0,0.5000]$ & $3.5077 \times 10^{0}$  & 2 \\
        0.5000 & 0.2500 & $[0,0.2500]$ & $2.6037 \times 10^{-4}$ & 4 \\
        0.2500 & 0.1250 & $[0,0.1250]$ & $1.2514 \times 10^{-4}$ & 8 \\
        0.1250 & 0.0625 & $[0,0.0625]$ & $2.5978 \times 10^{-4}$ & 16 \\
        \hline
    \end{tabular}
    \caption{Adaptive time interval partitioning for the 3D problem.}
    \label{Heat_adaTTTTTT}
\end{table}

\begin{table}[htbp]
    \centering
    \begin{tabular}{c|ccc}
        \hline
        Number of intervals  &  $\left\|\epsilon\right\|_{2}$  & $\|e\|_1 $ & $\|e\|_\infty$ \\ 
        \hline
1(Standard PINNs）& $6.4978 \times 10^{-1}$ & $8.9465 \times 10^{1}$ & $6.3867 \times 10^{2}$ \\
        4(THC-PINNs) & $1.7007 \times 10^{-3}$ & $1.4275 \times 10^{-1}$ & $1.2495 \times 10^{1}$ \\
        \hline
    \end{tabular}
    \caption{The computational errors with different PINN methods.}
    \label{Heat_fix_multgroup}
\end{table}

\begin{figure}[htbp]
\centering
\includegraphics[width=1\textwidth]{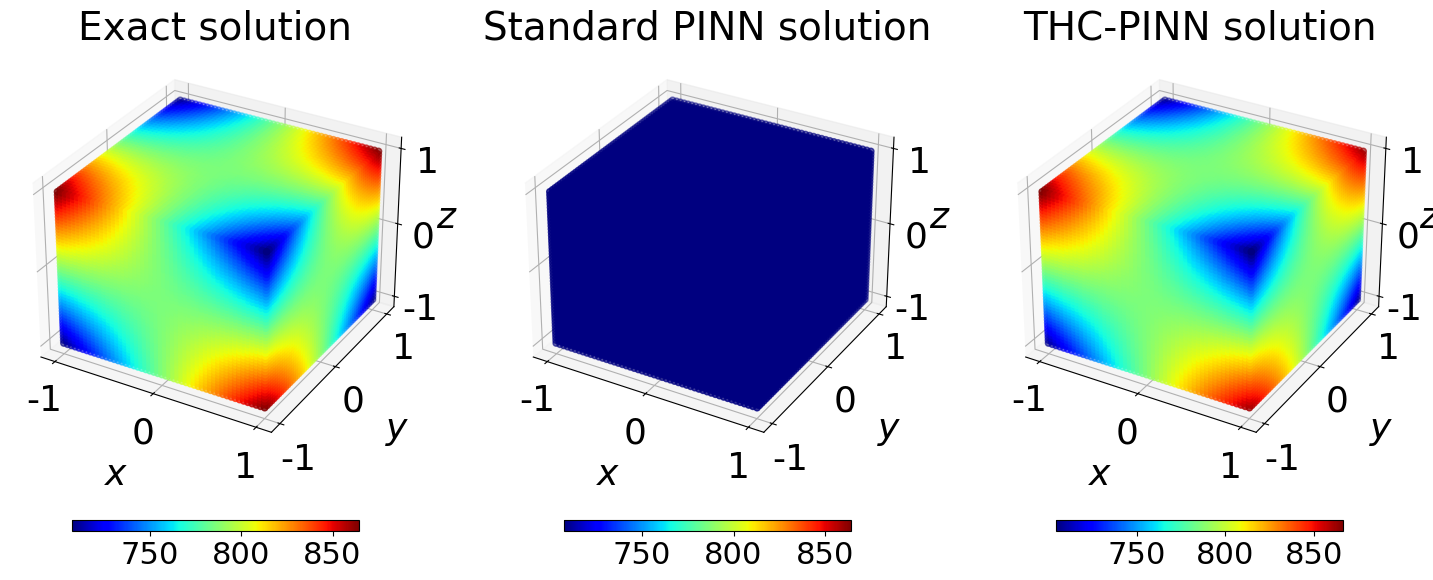}
\caption{Comparison of the PINN solutions   and the exact solution  at $t=1$.}
\label{Heat_tt}
\end{figure}

\begin{figure}[htbp]
\centering
\includegraphics[width=1\textwidth]{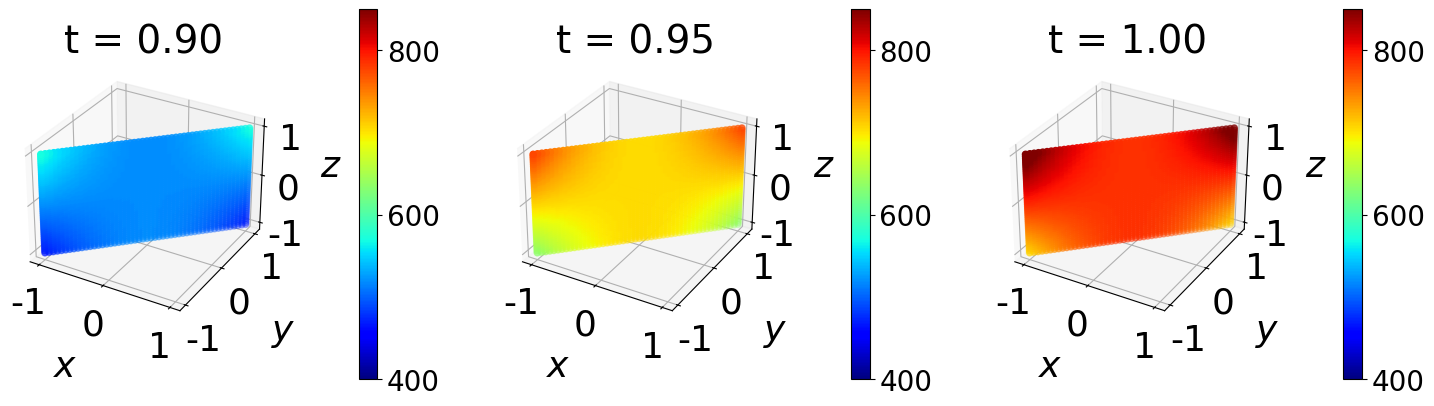}
\caption{THC-PINN solution  at  $x=y$ with different time.}
\label{Heat_xx}
\end{figure}

Table \ref{Heat_fix_multgroup} presents  the computational errors of standard PINNs and   THC-PINNs with four intervals. Figure \ref{Heat_tt} depicts  the   contours of exact solution,    standard PINN solution,  THC-PINN solution   at $t=1$.  
Figure \ref{Heat_xx} shows the slice of 
the THC-PINN solution on the face $x=y$ at  time  $t = 0.90, 0.95, 1.00$.
  These results show the superiority of our method for the time multi-scale problem.

\section{Conclusions}\label{sec5}

This paper introduces a deep learning method for evolutionary equations based on a novel hard constraint strategy. First, we utilize sequential learning strategies to divide the large temporal domain  into multiple intervals and train them sequentially, ensuring that the optimization process naturally  respects the principle of causality. Secondly, by designing a hard constraint strategy that includes influence functions, we ensure the continuity and smoothness of the PINN solution at interval nodes, while passing information from the previous time interval to the next, thereby avoiding incorrect or trivial solutions that deviate from the initial  interval. Furthermore, by introducing trainable parameters into the influence functions, we significantly enhance the method's generality across different types of equations, enabling it to adaptively solve various governing equations. 
This technique is particularly crucial for hyperbolic equations, as it plays an important role in improving the computational accuracy for solving such equations.
Additionally, we provide an adaptive temporal domain partitioning strategy, offering an effective approach to the reasonable selection of time domain partitioning. This strategy enhances the computational efficiency and accuracy of the overall problem by automatically selecting appropriate interval lengths, thereby avoiding issues such as training failure, reduced precision, or significantly increased training times due to intervals being too large or too small. Numerical experiments have validated the superior performance of the proposed method on several typical evolutionary equations.

\section*{Acknowledgements}
The work of Y. Yao is supported by the National Natural Science Foundation of China (No. 12271055,12471366) and National Key Research and Development Program of China (No. 2024YFA1012102). The work of S. Su is partially supported by the National Natural Science Foundation of China (No. 12201020).

\bibliography{HCPINN}

\begin{thebibliography}{10}
\expandafter\ifx\csname url\endcsname\relax
  \def\url#1{\texttt{#1}}\fi
\expandafter\ifx\csname urlprefix\endcsname\relax\def\urlprefix{URL }\fi
\expandafter\ifx\csname href\endcsname\relax
  \def\href#1#2{#2} \def\path#1{#1}\fi

\bibitem{712178}
I.~Lagaris, A.~Likas, D.~Fotiadis, Artificial neural networks for solving ordinary and partial differential equations, IEEE Transactions on Neural Networks 9~(5) (1998) 987--1000.

\bibitem{RAISSI2019686}
M.~Raissi, P.~Perdikaris, G.~Karniadakis, Physics-informed neural networks: A deep learning framework for solving forward and inverse problems involving nonlinear partial differential equations, Journal of Computational Physics 378 (2019) 686--707.

\bibitem{Karniadakis}
G.~E. Karniadakis, I.~G. Kevrekidis, L.~Lu, et~al., Physics-informed machine learning, Nature Reviews Physics 3 (2021) 422--440.

\bibitem{SOIBAM2024125480}
J.~Soibam, I.~Aslanidou, K.~Kyprianidis, R.~B. Fdhila, Inverse flow prediction using ensemble pinns and uncertainty quantification, International Journal of Heat and Mass Transfer 226 (2024) 125480.

\bibitem{PENWARDEN2023112464}
M.~Penwarden, A.~D. Jagtap, S.~Zhe, et~al., A unified scalable framework for causal sweeping strategies for physics-informed neural networks (pinns) and their temporal decompositions, Journal of Computational Physics 493 (2023) 112464.

\bibitem{Wang2020OnTE}
S.~Wang, H.~Wang, P.~Perdikaris, On the eigenvector bias of fourier feature networks: From regression to solving multi-scale pdes with physics-informed neural networks, Computer Methods in Applied Mechanics and Engineering 384 (2021) 113938.

\bibitem{WANG2024113112}
Y.~Wang, Y.~Yao, J.~Guo, Z.~Gao, A practical pinn framework for multi-scale problems with multi-magnitude loss terms, Journal of Computational Physics 510 (2024) 113112.

\bibitem{Wang2024RespectingCF}
S.~Wang, S.~Sankaran, P.~Perdikaris, Respecting causality for training physics-informed neural networks, Computer Methods in Applied Mechanics and Engineering 421 (2024) 116813.

\bibitem{WANG2024106998}
Y.~Wang, Y.~Yao, Z.~Gao, An extrapolation-driven network architecture for physics-informed deep learning, Neural Networks (2024) 106998.

\bibitem{CiCP-29-930}
C.~L.~Wight, J.~Zhao, Solving allen-cahn and cahn-hilliard equations using the adaptive physics informed neural networks, Communications in Computational Physics 29~(3) (2021) 930--954.

\bibitem{MATTEY2022114474}
R.~Mattey, S.~Ghosh, A novel sequential method to train physics informed neural networks for allen cahn and cahn hilliard equations, Computer Methods in Applied Mechanics and Engineering 390 (2022) 114474.

\bibitem{guo2023pretraining}
J.~Guo, Y.~Yao, H.~Wang, T.~Gu, Pre-training strategy for solving evolution equations based on physics-informed neural networks, Journal of Computational Physics 489 (2023) 112258.

\bibitem{JUNG2024117036}
J.~Jung, H.~Kim, H.~Shin, M.~Choi, Ceens: Causality-enforced evolutional networks for solving time-dependent partial differential equations, Computer Methods in Applied Mechanics and Engineering 427 (2024) 117036.

\bibitem{Exact}
P.~Roy, S.~Castonguay, Exact enforcement of temporal continuity in sequential physics-informed neural networks, Computer Methods in Applied Mechanics and Engineering 430 (2024) 117197.

\bibitem{RAISSI1}
M.~Raissi, P.~Perdikaris, G.~Karniadakis, Physics-informed neural networks: A deep learning framework for solving forward and inverse problems involving nonlinear partial differential equations, Journal of Computational Physics 378 (2019) 686--707.

\bibitem{article}
Y.~Huang, Z.~Xu, C.~Qian, L.~Liu, Solving free-surface problems for non-shallow water using boundary and initial conditions-free physics-informed neural network (bif-pinn), Journal of Computational Physics 479 (2023) 112003.

\bibitem{Projection}
Y.~Chen, D.~Huang, D.~Zhang, J.~Zeng, N.~Wang, H.~Zhang, J.~Yan, Theory-guided hard constraint projection (hcp): a knowledge-based data-driven scientific machine learning method, Journal of Computational Physics 445 (2021) 110624.

\bibitem{PhyCRNet}
P.~Ren, C.~Rao, Y.~Liu, J.-X. Wang, H.~Sun, Phycrnet: Physics-informed convolutional-recurrent network for solving spatiotemporal pdes, Computer Methods in Applied Mechanics and Engineering 389 (2022) 114399.

\bibitem{haitsiukevich2023improved}
K.~Haitsiukevich, A.~Ilin, Improved training of physics-informed neural networks with model ensembles, in: 2023 International Joint Conference on Neural Networks (IJCNN), IEEE, 2023, pp. 1--8.

\bibitem{Mojgani2022LagrangianPA}
R.~Mojgani, M.~Balajewicz, P.~Hassanzadeh, Lagrangian {PINNs}: A causality-conforming solution to failure modes of physics-informed neural networks, ArXiv 2205.02902 (2022).

\bibitem{Krishnapriyan2021CharacterizingPF}
A.~Krishnapriyan, A.~Gholami, S.~Zhe, R.~Kirby, M.~W. Mahoney, Characterizing possible failure modes in physics-informed neural networks, in: M.~Ranzato, A.~Beygelzimer, Y.~Dauphin, P.~Liang, J.~W. Vaughan (Eds.), Advances in Neural Information Processing Systems, Vol.~34, Curran Associates, Inc., 2021, pp. 26548--26560.

\bibitem{krishnapriyan2021characterizing}
A.~S. Krishnapriyan, A.~Gholami, S.~Zhe, et~al., Characterizing possible failure modes in physics-informed neural networks, Advances in Neural Information Processing Systems 34 (2021).

\bibitem{kingma2017adam}
D.~P. Kingma, J.~Ba, Adam: A method for stochastic optimization (2017).

\bibitem{LLLBFGS}
R.~H. Byrd, P.~Lu, J.~Nocedal, C.~Zhu, A limited memory algorithm for bound constrained optimization, SIAM Journal on Scientific Computing 16~(5) (1995) 1190--1208.

\bibitem{Vikhrenko_2011}
V.~S. Vikhrenko, Heat Conduction, IntechOpen, Rijeka, 2011.

\end{thebibliography}

\end{CJK}
\end{document}